\documentclass[journal,twocolumn,10pt]{IEEEtran}

\usepackage[cmex10]{amsmath}
\usepackage{amsfonts,amssymb}  
\usepackage{subfigure}
\usepackage{algorithm}
\usepackage{algpseudocode}
\usepackage{mathtools}
\usepackage{graphicx,subfigure}
\usepackage[usenames,dvipsnames]{color}
\usepackage[nospace]{cite}

\newcommand{\JS}[1]{{ \color{black}  #1 } } 

\newcommand{\argmin}{\operatornamewithlimits{argmin}}

\begin{document}
%
\title{Iterative Methods for Efficient Sampling-Based Optimal Motion Planning of Nonlinear Systems}
%
%
%

\author{Jung-Su Ha, Han-Lim Choi, and Jeong hwan Jeon
\thanks{J.-S. Ha and H.-L. Choi are  with the Department  of Aerospace Engineering, KAIST, Daejeon, Korea.     \texttt{\{wjdtn1404, hanlimc\}@kaist.ac.kr}}%
\thanks{J. Jeon  is with nuTonomy Inc., Cambridge, MA 02142, USA. \texttt{ jhjeon@alum.mit.edu}}
}

\maketitle

\begin{abstract}
This paper extends the RRT* algorithm, a recently-developed but widely-used sampling-based optimal motion planner, in order to effectively handle nonlinear kinodynamic constraints. Nonlinearity in kinodynamic differential constraints often leads to difficulties in choosing appropriate distance metric and in computing optimized trajectory segments in tree construction.  To tackle these two difficulties, this work adopts the affine quadratic regulator-based pseudo metric as the distance measure and utilizes iterative two-point boundary value problem solvers for computing the optimized segments. The proposed extension then preserves the inherent asymptotic optimality of the RRT* framework, while efficiently handling a variety of kinodynamic constraints. Three numerical case studies validate the applicability of the proposed method.
\end{abstract}


\IEEEpeerreviewmaketitle

\section{Introduction}

Robotic motion planning designs a trajectory of robot states from a given initial state to a specified goal state through a complex configuration space. While motion planning algorithms can be categorized into two groups: combinatorial and sampling-based approach~\cite{lavalle2011motion}, the latter (such as Probabilistic Road Map (PRM) and Rapidly-exploring Random Tree (RRT)) has been successful in practice as their computational advantage over combinatorial methods allows for handling of complex planning environments.
In particular, the Rapidly-exploring Random Tree Star (RRT*) algorithm proposed in \cite{Karaman11J} is one of the most influential algorithms of this type, as it guarantees probabilistic completeness and asymptotic optimality at the same time. In other words, RRT* guarantees that if the planning problem is feasible, the probability of the algorithm failing to find a solution reduces to zero as the number of iterations increase, and that the solution asymptotically approaches to the optimal solution; it is an ``anytime" algorithm which finds a feasible trajectory quickly and refines the solution for allowed computation time.
In addition, RRT* inherits the key advantage of RRT: it explores the unexplored search space rapidly \cite{lavalle1998rapidly}.
Due to these advantages, the algorithm has been successfully extended to many applications such as differential game and stochastic optimal control problems~\cite{karaman2011incremental,huynh2012incremental,huynh2014martingale}.

The motion planning problem is called an \textit{optimal kinodynamic} motion planning when the objective of planning is to minimize a given cost function defined by the state and control trajectory under system dynamics constraints. \JS{Applications of such problem include the automatic car-parking \cite{kim2010practical} and the trajectory planning of underwater vehicles \cite{yuan2009optimal} or gantry cranes\cite{blajer2007motion} in cluttered environment. The problem is challenging because the resulting trajectory should not only satisfies system dynamics but also lies over highly non-convex state space because of the obstacle field.} Due to the aforementioned properties, RRT* can provide a good framework for an optimal kinodynamic motion planner, supposed that the following two issues are appropriately addressed. First, the distance metric should be able to take into account kinodynamic constraints of the problem. RRT-based algorithms take advantage of Voronoi bias for rapid exploration of the state space; with a wrong distance metric, the configuration space may not be effectively explored. Second, there should be a way to construct a optimal trajectory segment under kinodynamic constraints for a given cost form, because the RRT* algorithm improves the quality of solution by refining the segments of the trajectory so that the solution asymptotically converges to the optimal solution.

There have been many attempts to handle the kinodynamic planning problem in the framework of RRT* by tackling the aforementioned two issues in some ways.
The minimum-time/length planning for holonomic and non-holonomic vehicles\cite{Karaman10, Karaman11, karaman2013sampling}  have first been addressed in the RRT* framework; a method tailored to high-speed off-road vehicles taking tight turns~\cite{Jeon11} was also proposed. Several recent researches have been devoted to deal with kinodynamic constraints in the form of linear differential constraint~\cite{Dustin13,perez2012lqr,goretkinoptimal}; these work in particular proposed to adopt the optimal control theory for linear systems for cost functions of some linear-\cite{perez2012lqr,goretkinoptimal} or affine-quadratic regulator\cite{Dustin13} (LQR or AQR) type. Despite these recent progresses, the question of a systematic and efficient method to handle generic nonlinear dynamics, which inevitably involves computation of two-point boundary value problems (TPBVPs) solutions in the RRT* process remains unsettled.

This work focuses on presenting methodology that can effectively handle nonlinear dynamics in the framework RRT*. The methodology finds out an optimal trajectory for an affine-quadratic cost functional under nonlinear differential constraints while allowing for rapid exploration of the state space.  The AQR-based pseudo metric is proposed as an approximation to the optimum distance under nonlinear differential constraints, and two iterative methods are presented to solve associated TPBVPs efficiently. The proposed extension of RRT* preserves asymptotic optimality of the original RRT*, while taking into account a variety of kinodynamic constraints. Three numerical case studies are presented to demonstrate the applicability of the proposed methodology.

While one of the two iterative methods in this paper was first introduced in the authors' earlier work~\cite{ha2013sa}, this article includes more extended description of the methodology, in particular proposing one more iterative algorithm, as well as more diverse/extensive numerical case studies.

\section{Problem Definition}
A kinodynamic motion planning problem is defined for a dynamical system
\begin{equation}
\dot{x}(t)=f(x(t),u(t)), \label{eq:dyn_nonl}
\end{equation}
where $x$ denotes the state of the system defined over the state space $\chi \subset \mathbb{R}^n$ and $u$ denotes the control input defined over the control input space $U \subset \mathbb{R}^m$.
Let $\chi_{obs} \subset \chi$ and $\chi_{goal} \subset \chi$ be the obstacle region and the goal region where the system tries to avoid and to reach, respectively.
Then,  the feasible state and input spaces are given by $\chi_{free} \subset \chi\setminus\chi_{obs}$ and $U_{free} \subset U$, respectively. 

The trajectory is represented as $\pi = (x(\cdot),u(\cdot), \tau)$, where $\tau$ is the arrival time at the goal region, $u:[0,\tau]\rightarrow U$, $x:[0,\tau]\rightarrow\chi$ are the control input and the corresponding state along the trajectory. 
The trajectory, $\pi_{free}$, for given initial state $x_{init}$ is called feasible if it does not cross the obstacle region and eventually achieves the goal region while satisfying the system dynamics (\ref{eq:dyn_nonl}), i.e., $\pi_{free} = (x(\cdot),u(\cdot), \tau)$, where $u:[0,\tau]\rightarrow U_{free}$, $x:[0,\tau]\rightarrow\chi_{free}$,  $x(0) = x_{init}$ and $x(\tau) \in \chi_{goal}$.

In order to evaluate a given trajectory $\pi$, the below form of the cost functional is considered in this paper:
\begin{equation}
c(\pi) =\int^{\tau}_{0} \left[1+\frac{1}{2}u(t)^TRu(t) \right]dt. \label{eq:cost}
\end{equation}
The above cost functional denotes trade-off between arrival time of a trajectory and the expanded control effort.
$R$ is user-define value; the cost function penalizes more for the trajectory spending large control effort than late arrival time as $R$ is larger.
This type of cost functional is widely used for the kinodynamic planning problems \cite{Glassman10, Dustin13}.

Then finally, the problem is defined as follows.\\
\textbf{Problem 1 (Optimal kinodynamic motion planning):} \textit{Given $\chi_{free},~x_{init}$ and $\chi_{goal}$, find a minimum cost trajectory $\pi^* = (x^*(\cdot),u^*(\cdot), \tau^*)$ such that $\pi^* = \argmin_{\pi\in\Pi_{free}} c(\pi)$, where $\Pi_{free}$ denotes a set of feasible paths.}

\section{Background}  \label{sec:background}
\subsection{The RRT* Algorithm}
This section summarizes the RRT* algorithm~ \cite{Karaman11J} upon which this paper builds an extension to deal with kinodynamic constraints.  The RRT* is a sampling-based algorithm that incrementally builds and produces an optimal trajectory from a specified initial state $x_{init}$ to a specified goal region $X_{goal}$. The overall structure of the algorithm is summarized in Algorithm \ref{RRT*}. It should be pointed out that Algorithm \ref{RRT*} uses more generic notations/terminologies than the original algorithm to better describe the proposed kinodynamic extension within the presented algorithm structure.

\begin{algorithm}
\caption{RRT* algorithm}\label{RRT*}
\begin{algorithmic}[1]
\State $(V,E) \gets (\{x_{init}\},\emptyset);$
\For{$i=1,...,N$}
\State $x_{rand}\gets \textsc{Sampling}(\chi_{free});$
\State $x_{nearest}\gets \textsc{Nearest}(V, x_{rand});$
\State $x_{new}\gets \textsc{Steer}(x_{nearest}, x_{rand});$
\State $\pi_{new}\gets\textsc{TPBVPsolver}(x_{nearest},x_{new}); $
\If{$\textsc{ObstacleFree}(\pi_{new})$}
\State $X_{near\_b}\gets \textsc{NearBackward}(V,x_{new});$
\State $X_{near\_f}\gets \textsc{NearForward}(V,x_{new});$
\State $c_{min}\gets \textsc{Cost}(x_{nearest})+c(\pi_{new});$
\For{$x_{near}\in X_{near\_b}$} 
\State $\pi_{new}'\gets \textsc{TPBVPsolver}(x_{near},x_{new});$
\State $c'\gets \textsc{Cost}(x_{near})+c(\pi_{new}');$
\If{$\textsc{ObstacleFree}(\pi_{new}')\wedge c'<c_{min}$}
\State $c_{min}\gets c';~\pi_{new}\gets \pi_{new}';$
\EndIf
\EndFor
\State $V\gets V \cup x_{new};~E\gets E\cup\pi_{new};$
\For{$x_{near}\in X_{near\_f}$} 
\State $\pi_{near}'\gets \textsc{TPBVPsolver}(x_{new},x_{near});$
\State $c'\gets \textsc{Cost}(x_{new})+c(\pi_{near}');$
\If{$\textsc{ObstacleFree}(\pi_{near}')\wedge c'<\textsc{Cost}(x_{near})$}
\State $E\gets (E \setminus \pi_{near}) \cup \pi_{near}';$ \Comment replace existing edge
\EndIf
\EndFor
\EndIf
\EndFor
\State \Return $T\gets (V,E);$
\end{algorithmic}
\end{algorithm}

At each iteration, the algorithm randomly samples a state $x_{rand}$ from $\chi_{free}$;  finds the \textit{nearest} node $x_{nearest}$ in the tree to this sampled state (line 3-4).
Then, the algorithm steers the system toward $x_{rand}$ to determine $x_{new}$ that is closest to $x_{rand}$ and stays within some specified distance from $x_{nearest}$; then, add $x_{new}$ to the set of vertices $V$ if the trajectory from $x_{nearest}$ to $x_{new}$ is obstacle-free (lines 5 -- 7 and 18).
Next, the best parent node for $x_{new}$  is chosen from near nodes in the tree so that the trajectory from the parent to $x_{new}$ is obstacle-free and of minimum cost (lines 8, 10 -- 17). 
After adding the trajectory segment from the parent to $x_{new}$ (line 18), the algorithm rewires the near nodes in the tree so that the forward paths from  $x_{new}$ are of minimum cost (lines 9, 19 -- 25); 
Then, the algorithm proceeds to the next iteration.

To provide more detailed descriptions of the key functions of the algorithm:
\begin{itemize}
\item \textsc{Sampling}($\chi_{free})$: randomly samples a state from $\chi_{free}$.
\item \textsc{Nearest}($V,x$): finds the nearest node from $x$ among nodes in the tree $V$ under a given distance metric \textit{dist$(x_1,x_2)$} representing distance from $x_1$ to $x_2$.
\item \textsc{Steer}($x,y$): returns a new state $z\in\chi$ such that $z$ is closest to $y$ among candidates:
\begin{equation}
\textsc{Steer}(x,y) := \argmin_{z\in B^+_{x,\eta}}\textit{dist}(z, y), \nonumber
\end{equation}
where $B^+_{x,\eta}\equiv\{z\in\chi|\textit{dist(x, z)}\leq\eta\}$ with $\eta$ representing the maximum length of an one-step trajectory forward.
\item \textsc{TPBVPsolver($x_0,x_1$):} returns the optimal trajectory from $x_0$ to $x_1$, \textit{without} considering the obstacles.
\item \textsc{ObstacleFree($\pi$)}: returns indication of whether or not the trajectory $\pi$ overlaps with the obstacle region $\chi_{obs}$.
\item \textsc{NearBackward($V,x$)} and \textsc{NearForward($V,x$)}: return the set of nodes in $V$ that are within the distance of $r_{|V|}$ from/to $x$, respectively.
In other words,
\begin{equation}
\textsc{NearBackward}(V, x) := \{v\in V|v\in B^-_{x,r_{|V|}}\}, \nonumber
\end{equation}
\begin{equation}
\textsc{NearForward}(V, x) := \{v\in V|v\in B^+_{x,r_{|V|}}\}, \nonumber
\end{equation}
where $B^-_{x,r_{|V|}}\equiv\{z\in\chi|\textit{dist}(z, x)\leq r_{|V|}\}$, $B^+_{x,r_{|V|}}\equiv\{z\in\chi|\textit{dist}(x, z)\leq r_{|V|}\}$, and $|V|$ denotes the number of nodes in the tree.

Also, $r_{|V|}$ needs to be chosen such that a ball of  volume $\gamma\frac{\log |V|}{|V|}$ is contained by $B^-_{x,r_{|V|}}$ and  $B^+_{x,r_{|V|}}$ with large enough $\gamma$.
For example, with the Euclidean distance metric  this can be defined as $r_{|V|}=\min\{(\frac{\gamma}{\zeta_d}\frac{\log n}{n})^{1/d},\eta\}$  with the constant $\gamma$~\cite{Karaman11J}; $\zeta_d$ is the volume of the unit ball in $\mathbb{R}^d$.
\item  $\textsc{Cost}(x)$: returns the cost-to-come for node $x$ from the initial state.
\item $\textsc{Parent}(x)$: returns a pointer to the parent node of $x$.
\item $c(\pi)$: returns the cost of trajectory $\pi$ defined by (\ref{eq:cost}).
\end{itemize}

Note specifically that  when the cost of a trajectory is given by the path length and \text{no} kinodynamic constraint is involved (as was in the first version presented in \cite{Karaman11J}),  the Euclidean distance can be used as the distance metric $dist(x_1,x_2)$. Thus,  $\textsc{NearBackward}(V,x)$ becomes identical to $\textit{NearForward}(V,x)$ because $dist(x,z)  = dist(z,x)$, and  $\textsc{TPBVPsolver}(x_0,x_1)$ simply returns a straight line from $x_0$ to $x_1$.


The probabilistic completeness and asymptotic optimality of RRT* algorithm are proven in \cite{Karaman11J}: if the planning problem has at least one feasible solution, the probability that the algorithm cannot finds the  solution goes to zero as the number of iterations increases, and at the same time, the solution is being refined to the optimal one asymptotically.

It should be noted that  the RRT* structure in Algorithm \ref{RRT*} can be adopted/extended for \textit{kinodynamic} optimal planning, supposed that the distance metric takes into account the kinodynamic constraints and that the \textsc{TPBVPsolver} computes the optimal trajectory between two states under differential kinodynamic constraints.

\subsection{Optimal Control with Affine Dynamics}\label{subsec:OC}
This section presents a procedure to compute the optimal solution for an affine system with affine-quadratic cost functional, which will be taken advantage of for quantification of the distance metric for generic nonlinear systems later in the paper. Consider an affine system
\begin{equation}
\dot{x}(t) = Ax(t)+Bu(t)+c,\label{eq:dyn_l}
\end{equation}
and the performance index
\begin{equation}
J(0) = \int_{0}^{\tau} \left[1+\frac{1}{2}u(t)^TRu(t) \right] dt,\label{eq:perf_ind}
\end{equation}
with boundary conditions,
\begin{equation}
x(0) = x_0, \qquad  x(\tau) = x_1 \label{eq:boundary_affine}
\end{equation}
and \textit{free} final time $\tau$.

The Hamiltonian is given by,
\begin{equation}
H = 1+\frac{1}{2}u(t)^TRu(t)+\lambda(t)^T(Ax(t)+Bu(t)+c). \nonumber
\end{equation}
The minimum principle yields that the optimal control takes the form of:
\begin{equation}
u(t)=-R^{-1}B^T\lambda(t), \nonumber
\end{equation}
with a reduced Hamilitonian system~\cite{lewis1995optimal}:
\begin{align}
\dot{x}(t) &= Ax(t)-BR^{-1}B^T\lambda(t)+c, \notag\\
-\dot{\lambda}(t) &= A^T\lambda(t),\label{eq:TPBVP1} \tag{\textbf{BVP1}}
\end{align}
where $x(0) = x_0,~x(\tau) = x_1$.
The solution of (\ref{eq:TPBVP1}) with the boundary conditions (\ref{eq:boundary_affine})  is the optimal trajectory from $x_0$ to $x_1$, supposed that the final time $\tau$ is given.

Note that for given $\tau$, the terminal values of $x(t)$ and $\lambda(t)$ in (\ref{eq:TPBVP1}) can be expressed:
\begin{equation}
x(\tau) = x_1, \lambda(\tau) = -G(\tau)^{-1}(x_1-x_h(\tau)), \label{eq:bc1}
\end{equation}
where a homogeneous solution of (\ref{eq:dyn_l}), $x_h(\tau)$, and the weighted continuous reachability Gramian, $G(\tau)$, are the final values of the following initial value problem: 
\begin{align}
\dot{x}_h(t) &= Ax_h(t)+c,\notag\\
\dot{G}(t) &= AG(t)+G(t)A^T+BR^{-1}B^T,\notag\\
x_h(0) &= x_0, G(0) = 0.\label{eq:IVP1} \tag{\textbf{IVP1}}
\end{align}
With (\ref{eq:bc1}), the optimal trajectory can be  obtained by integrating (\ref{eq:TPBVP1}) backward.

For given final time $\tau$, the performance index of the optimal trajectory can be written as
\begin{equation}
C(\tau) = \tau + \frac{1}{2}(x_1-x_h(\tau))^TG(\tau)^{-1}(x_1-x_h(\tau)). \label{eq:cost_tau}
\end{equation}
and the optimal final time can be expressed as:
\begin{equation}
\tau^* = \argmin_{\tau\geq0} C(\tau). \label{eq:opt_tau}
\end{equation}
Equation (\ref{eq:cost_tau}) implies that $C(\tau) \geq \tau$ since $G(\tau)$ is positive (semi-)definite.
Therefore, $\tau^*$ can be computed by calculating $C(\tau)$ with increasing $\tau$ until $\tau$ equals to the incumbent best cost $\widetilde{C}(\tau) \triangleq \min_{t \in [0,\tau]} C(t)$. With this optimal final time $\tau^*$, the optimal cost can be obtained as:
 \begin{equation}
 C^* = C(\tau^*). \label{eq:opt_cost}
 \end{equation}

Within the RRT* framework, if the kinodynamic constraints take an affine form, the procedure in this section can be used to define/quantify the distance metric and to compute the optimized trajectory between two nodes in $\textsc{TBPVPsolver}(x_0, x_1)$.

\section{Efficient Distance Metric and Steering Method}\label{sec:Dis}
To take advantage of the explorative property of RRT*, it is crucial to use an appropriate distance metric in the process. The Euclidean distance, which cannot consider the system dynamics, is certainly not a valid option in kinodynamic motion planning problem -- for example, it would fail to find the nearest node and thus would not be able to steer toward the sample node. Thus, a distance metric which appropriately represents the degree of closeness taking into account the dynamic constraints and the underlying cost measure needs to be defined/quantified for kinodynamic version of RRT*.

\subsection{Affine Quadratic Regulator-Based Pseudo Metric}
An AQR-based pseudo metric is firstly proposed in \cite{Glassman10} as a distance metric to kinodynamic planning to consider a first-order linear dynamics. In this work, an AQR-based pseudo metric is adopted as an approximate distance measure for problems with nonlinear differential constraints. For kinodynamic planning with cost functional (\ref{eq:cost}) and dynamic constraint (\ref{eq:dyn_nonl}), the distance from $x_0$ to $x_1$ is computed as
\begin{equation}
\textit{dist}(x_0, x_1) = C^* \equiv \min_{\tau\geq 0}C(\tau). \label{eq:dist}
\end{equation}
where $C(\tau)$ is calculated from (\ref{eq:cost_tau}) with
$$
A = \left.\frac{\partial f}{\partial x}\right|_{x=\hat{x}, u = \hat{u}}, ~~ B = \left.\frac{\partial f}{\partial u}\right|_{x=\hat{x}, u = \hat{u}},
$$
where $\hat{x}$ and $\hat{u}$ indicate the linearization points that are set to be the initial (or the final) state, i.e., $\hat{x} = x_0$ (or $x_1$), and $\hat{u}=0$, in the framework of RRT*.
In other words, the distance from $x_0$ to $x_1$ is approximated as the cost of the optimal control problem for a linearized system with the same cost functional and boundary conditions.

In this work, an AQR-based pseudo metric is adapted as a distance metric in the RRT* framework.
An exactness of a metric is related to the property of rapid exploration of the RRT* algorithm:
for a randomly chosen sample, the algorithm finds the nearest node by \textsc{Nearest} procedure and expands the tree toward the sample in \textsc{Steer} procedure.
Calculation of the exact metric between two states are equivalent to solving nonlinear optimal control problem which is computationally expensive;
the algorithm needs to solve such optimal control problems for every pair of states, from the sample state to the nodes in the tree and vice versa.
Although the AQR-based pseudo metric does not take into account the nonlinear dynamics, as will be mentioned in next subsection, it can measure the metric for all pairs of states by integrating only $n$ (for $x_h(t)$) + $n(n+1)/2$ (for $G(t)$ which is symmetric) first order ODEs and produces the much more exact degree of closeness than Euclidean distance.

\subsection{Implementation of AQR Metric in RRT* Framework}
Based on the AQR metric in (\ref{eq:dist}), \textsc{Nearest}($V,x$), \textsc{NearBackward}($V,x$), \textsc{NearForward}($V,x$) and \textsc{Steer}($x,y$) functions can be readily implemented in the RRT* structure.
\begin{itemize}
\item $x_{nearest}\gets \textsc{Nearest}(V, x_{rand})$: First, the system dynamics is linearized at $x_{rand}$.
Since $x_{rand}$ is final state, $x_h(t)$ and $G(t)$ are integrated from $t=0, x_h(0) = x_{rand}, G(0) = 0$ in the backward direction, i.e., $t<0$.
With integration, the cost from $i$th node $v_i \in V$ at the time $-t$ is calculated as $C_i(-t) = -t - \frac{1}{2}(v_i-x_h(t))^TG(t)^{-1}(v_i-x_h(t))$ while the minimum cost is saved as a distance, $d_i = \min_{t<0}C_i(-t)$.
The integration stops when $\min_{i}d_i\geq-t$; since $G(t)$ is a negative (semi-)definite matrix, cost less than $-t$ cannot be found by more integration.
Finally, the nearest node, $x_{nearest}\gets v_k$, where $k = \argmin_{i}d_i$, is returned.
\item $X_{near\_b}\gets \textsc{NearBackward}(V,x_{new})$: The procedure is similar to \textsc{Nearest}.
First, the system dynamics is linearized at $x_{new}$.
$x_h(t)$ and $G(t)$ are integrated from $t=0, x_h(0) = x_{new}, G(0) = 0$ for backward direction, i.e., $t<0$.
With integration, the cost for the time $-t$ and for each $v_i \in V$, the corresponding cost is calculated while the minimum cost is saved as a distance, $d_i = \min_{t<0}C_i(-t)$.
The integration stops when $-t\geq r_{|V|}$ and the set of backward near nodes, $X_{near\_b} \gets \{v_i\in V|d_i \leq r_{|V|}\}$, are returned.
\item $X_{near\_f}\gets \textsc{NearForward}(V,x_{new})$: The procedure is exactly same as \textsc{NearBackward} except for the direction of integration, $t>0$.
\item $x_{new}\gets \textsc{Steer}(x_{nearest}, x_{rand})$: First, the system dynamics is linearized at $x_{nearest}$.
Then, the optimal trajectory is calculated by procedure in section \ref{subsec:OC} for the linearized system.
If the resulting cost is less than $\eta$, $x_{rand}$ is returned as a new node.
Otherwise, it returns $x'$ such that $x'$ is in the trajectory and the cost to $x'$ from $x_{nearest}$ is $\eta$.
\end{itemize}

\section{Efficient Solver for the Two-Point Boundary-Value Problem}\label{sec:TPBVP}
As mentioned previously, the $\textsc{TPBVPsolver}(x_0,x_1)$ function returns the optimal trajectory from $x_0$ to $x_1$.
A straight-line trajectory, which is optimal for the problem without dynamic constraints, cannot be a valid solution in general, because it would be not only suboptimal but also likely to violate the kinodynamic constraints.  Therefore, this section derives the two-point boundary value problem (TPBVP) involving nonlinear differential constraints and presents methods to compute the solution of this TPBVP.

Consider the optimal control problem (OCP) with nonlinear system dynamics in (\ref{eq:dyn_nonl}) to minimize the cost functional in  (\ref{eq:perf_ind}) where boundary conditions are given as $ x(0) = x_0,~x(\tau) = x_1$ with free final time $\tau$.
The Hamiltonian of this OCP is defined as,
\begin{equation}
H(x(t),u(t),\lambda(t)) = 1+\frac{1}{2}u(t)^TRu(t)+\lambda(t)^Tf(x(t),u(t)). \nonumber
\end{equation}
From the minimum principle:
\begin{equation}
H_u^T = Ru(t)+f_u^T\lambda(t) = 0,\nonumber
\end{equation}
which allows the optimal control to be expressed in terms of $x(t)$ and $\lambda(t)$:
\begin{equation}
u(t) = h(x(t),\lambda(t)) \label{eq:con}.
\end{equation}
Thus, a system of differential equations for the state $x(t)$ and the costate $\lambda(t)$ is obtained:
\begin{align}
\dot{x}(t) &= f(x(t), h(x(t),\lambda(t))), \notag\\
-\dot{\lambda}(t) &= H_x^T = f_x^T\lambda(t)
\label{eq:TPBVP2} \tag{\textbf{BVP2}}
\end{align}
with boundary conditions $ x(0) = x_0,~x(\tau) = x_1$.
The system of differential equations in (\ref{eq:TPBVP2}) is nonlinear in general and has boundary conditions at the initial and final time; thus, it is called a nonlinear two-point boundary value problem (TPBVP). An analytic solution to a nonlinear TPBVP is generally unavailable because not only  it is nonlinear but also its the boundary conditions are split at two time instances; numerical solution schemes have often been adopted for this.

In the present work, two types of numerical \textit{iterative} approaches are presented: the successive approximation (SA) and the variation of extremals (VE) based methods.
These two approaches find the solution of a nonlinear TPBVP by successively solving a sequence of more tractable problems: in the SA based method, a sequence of  linear TPBVPs are solved and in the VE approach, a sequence of nonlinear initial value problems (IVPs) are iteratively solved, respectively.
The main concept of the methods has already been presented to solve a \textit{free final-state \& fixed final-time} problem \cite{Tang05, kirk2012optimal} but this work proposes their variants that can handle  a \textit{fixed final-state \& free final-time} problem in order to implemented to the RRT* framework.

In an iterative method, an initial guess of the solution is necessary and often substantially affects the convergence of the solution. In the proposed RRT* extension, the optimal trajectory with the linearized dynamics can be a good choice of initial condition, particularly because such trajectory is already available in the RRT* process by calculating the AQR-based distance metric in \textsc{NearBackward} or \textsc{NearForward}.

\subsection{Successive Approximation}\label{sec:SA}
Let $\hat{x}$ be a linearization point, which is set to be the initial or the final value of a trajectory segment in the RRT* implementation; $\hat{u}=0$ be also the corresponding linearization point.
By splitting the dynamic equation in (\ref{eq:dyn_nonl}) into the linearized and remaining parts around $\hat{x}$ and $\hat{u}$,
$$
\dot{x}(t) = Ax(t)+Bu(t)+g(x(t),u(t)),
$$
where $A \triangleq \left.\frac{\partial f}{\partial x}\right|_{x=\hat{x}, u = \hat{u}}$, $B \triangleq \left. \frac{\partial f}{\partial u}\right|_{x=\hat{x}, u = \hat{u}}$ and $g(x(t),u(t))\triangleq f(x(t),u(t))-Ax(t)-Bu(t)$,
the optimal control and the reduced Hamiltonian system are expressed as,
\begin{equation}
u(t) = -R^{-1}B^T\lambda(t)-R^{-1}g_u^T\lambda(t),
\end{equation}
and
\begin{align}
\dot{x}(t) &= Ax(t)-BR^{-1}B^T\lambda(t)-BR^{-1}g_u^T\lambda(t) \nonumber\\
&~~~   +g(x(t),u(\lambda (t))),\nonumber\\
-\dot{\lambda}(t) &= A^T\lambda(t)+g_x^T\lambda(t), \label{eq:tpbvp_sa}
\end{align}
respectively.
The boundary conditions are given by $$x(0) = x_0, x(\tau) = x_1$$ and the final time $\tau$ is \textit{free}.
The TPBVP in (\ref{eq:tpbvp_sa}) is still nonlinear, as $g_u^T\lambda$, $g$ and $g_x^T\lambda$ in (\ref{eq:tpbvp_sa}) are \textit{not} necessarily linear w.r.t $x$ or $\lambda$.

Consider the following sequence of TPBVPs:
\begin{align}
\dot{x}^{(k)}(t) &= Ax^{(k)}(t)-BR^{-1}B^T\lambda^{(k)}(t) \nonumber\\
& ~~~-BR^{-1}g_u^{(k-1)T}\lambda^{(k-1)}(t)+g^{(k-1)}, \label{eq:x_seq}\\
-\dot{\lambda}^{(k)}(t) &= A^T\lambda^{(k)}(t)+g_x^{(k-1)T}\lambda^{(k-1)}(t), \label{eq:lamb_seq}
\end{align}
with boundary conditions $x^{(k)}(0)=x_0,~x^{(k)}(\tau)=x_1$, where $g^{(k)}\equiv g(x^{(k)},u^{(k)}),~g_x^{(k)}\equiv g_x(x^{(k)},u^{(k)})$ and $g_u^{(k)} \equiv g_u(x^{(k)},u^{(k)})$.
Note that this system of differential equations are linear w.r.t. $x^{(k)}(t)$ and $\lambda^{(k)}(t)$ for given $x^{(k-1)}(t)$ and $\lambda^{(k-1)}(t)$. It can be converted into an initial value problem as follows.  From (\ref{eq:lamb_seq}), $\lambda^{(k)}(t)$ can be expressed as:
\begin{equation}
\lambda^{(k)}(t) = e^{A^T(\tau-t)}\lambda^{(k)}(\tau)+\lambda_p^{(k)}(t), \label{eq:alg1}
\end{equation}
where $\lambda_p^{(k)}(t)$ is a solution of
\begin{equation}
\dot{\lambda}_p^{(k)}(t) = -A^T\lambda_p^{(k)}(t)-g_x^{(k-1)T}\lambda^{(k-1)}(t), \label{eq:lp}
\end{equation}
with terminal condition $\lambda_p^{(k)}(\tau) = 0$.
Plugging (\ref{eq:alg1}) into (\ref{eq:x_seq}) yields
\begin{align}
\dot{x}^{(k)}(t) &= Ax^{(k)}(t)+g^{(k-1)}-BR^{-1}B^Te^{A^T(T-t)}\lambda(\tau)\nonumber\\
&~~-BR^{-1}B^T\lambda_p^{(k)}(t)-BR^{-1}g_u^{(k-1)T}\lambda^{(k-1)}(t).
\end{align}
Then, the final state $x^{(k)}(\tau)$ is expressed as,
\begin{equation}
x^{(k)}(\tau) = x^{(k)}_h(\tau)-G(\tau)\lambda^{(k)}(\tau), \label{eq:xf_seq}
\end{equation}
where $x_h(\tau)$ and $G(\tau)$ are the solution of
\begin{align}
\dot{x_h}^{(k)}(t) &= Ax^{(k)}_h(t)-BR^{-1}B^T\lambda_p^{(k)}(t)\nonumber\\
&~~~-BR^{-1}g_u^{(k-1)T}\lambda^{(k-1)}(t)+g^{(k-1)} \label{eq:sa1}\\
\dot{G}(t) &= AG(t)+G(t)A^T+BR^{-1}B^T,\label{eq:sa2}\\
x^{(k)}_h(0) &= x_0, G(0) = 0. \nonumber
\end{align}
Using (\ref{eq:xf_seq}) and $x^{(k)}(\tau) = x_1$, the final costate value can be obtained:
\begin{equation}
\lambda^{(k)}(\tau)=-G(\tau)^{-1}(x_1-x^{(k)}_h(\tau)).
\end{equation}
Finally, $x^{(k)}(t)$ and $\lambda^{(k)}(t)$ for $t\in[0,\tau]$ are calculated by backward integration of the following differential equation,
\begin{align}
\begin{bmatrix}\dot{x}^{(k)}(t)\\ \dot{\lambda^{(k)}}(t) \end{bmatrix} =& \begin{bmatrix}A & -BR^{-1}B^T\\ 0 & -A^T \end{bmatrix}\begin{bmatrix}x^{(k)}(t)\\ \lambda^{(k)}(t)\end{bmatrix} \label{eq:alg2}\\
&+\begin{bmatrix}-BR^{-1}g_u^{(k-1)T}\lambda^{(k-1)}(t)+g^{(k-1)}\\ -g_x^{(k-1)T}\lambda^{(k-1)}(t)\end{bmatrix}, \nonumber
\end{align}
with given boundary values  $x^{(k)}(\tau)$ and $\lambda^{(k)}(\tau) $.
Also, the optimal control $u^{(k)}(t)$ can also be computed accordingly:
\begin{equation}
u^{(k)}(t) = -R^{-1}B^T\lambda^{(k)}(t)-R^{-1}g_u^{(k-1)T}\lambda^{(k-1)}(t). \label{eq:con_seq}
\end{equation}

The optimal final time can be found numerically by a gradient descent scheme.
The derivative of $J(0)$ for the trajectory at $k$'th iteration w.r.t final time is given as,
\begin{align}
\left[ \frac{dJ(0)}{d\tau}\right]^{(k)} &= 1-\frac{1}{2}\lambda^{(k)}(\tau)^TBR^{-1}B^T\lambda^{(k)}(\tau)\nonumber\\
&~~+\frac{1}{2}\lambda^{(k-1)}(\tau)^Tg_u^{(k-1)}R^{-1}g_u^{(k-1)T}\lambda^{(k-1)}(\tau)\nonumber\\
&~~+\lambda^{(k)}(\tau)^T(Ax_1+g^{(k)}). \label{eq:tau_seq1}
\end{align}
Then, a gradient descent-based update rule of the final time is given as,
\begin{equation}
\tau^{(k+1)} = \tau^{(k)} - \eta \left[\frac{dJ(0)}{d\tau} \right]^{(k)}. \label{eq:tau_seq2}
\end{equation}

With the process described thus far, the TPBVP solver for nonlinear system based on successive-approximation can be summarized as Algorithm \ref{alg:SA}.
\begin{algorithm}
\caption{$\pi\gets$SA-based TPBVPsolver$(x_0,x_1)$}\label{alg:SA}
\begin{algorithmic}[1]
\State Initialize $x^{(0)},~\lambda^{(0)},~\tau^{(0)}$ and set $k=0$ \Comment{Section \ref{subsec:OC}}
\Repeat
\State $k = k+1;$
\State Update $\tau^{(k)}$ \Comment{Equation (\ref{eq:tau_seq1})-(\ref{eq:tau_seq2})}
\State Get $x^{(k)}(t)$ and $\lambda^{(k)}(t)$\Comment{Equation (\ref{eq:alg1})-(\ref{eq:alg2})}
\Until{Converge}
\State Calculate $u(t)$ \Comment{Equation (\ref{eq:con_seq})}
\State \textbf{return} $\pi \gets (x^{(k)}(\cdot),u(\cdot), \tau^{(k)})$
\end{algorithmic}
\end{algorithm}

\subsection{Variation of Extremals}\label{sec:VarofEx}
The VE approach is a technique to successively find the initial value of the costate $\lambda(0)$ using the Newton-Rapson method \cite{kirk2012optimal}. The method has been well established, but its implementation into a sampling-based planning is not trivial.

Suppose that the initial value of the costate $\lambda(0)$ and the final value of the state $x(\tau)$ are related via nonlinear function:
\begin{equation}
x(\tau)=F(\lambda(0)). \nonumber
\end{equation}
Then, the  initial costate $\lambda(0)$ that leads to $x(\tau) = x_1$ can be approximated successively as follows:
\begin{equation}
\lambda^{(k+1)}(0) = \lambda^{(k)}(0) - [P_x(\lambda^{(k)}(0),\tau)]^{-1}(x^{(k)}(\tau)-x_1), \nonumber
\end{equation}
where $P_x(p^{(k)}(0),t)$ is called the \textit{state influence function matrix} which is the matrix of partial derivatives of the components of $x(t)$, evaluated at $p^{(k)}(0)$,
\begin{equation}
P_x(\lambda^{(k)}(0),t)\equiv \begin{bmatrix}\frac{\partial x_1(t)}{\partial \lambda_1(0)} & \cdots & \frac{\partial x_1(t)}{\partial \lambda_n(0)} \\
\vdots & \ddots & \vdots \\ \frac{\partial x_n(t)}{\partial \lambda_1(0)} & \dots & \frac{\partial x_n(t)}{\partial \lambda_n(0)}\end{bmatrix}_{\lambda^{(k)}(0)}. \nonumber
\end{equation}
Similarly, the \textit{costate influence function matrix} is defined and given by,
\begin{equation}
P_\lambda(\lambda^{(k)}(0),t)\equiv \begin{bmatrix}\frac{\partial \lambda_1(t)}{\partial \lambda_1(0)} & \cdots & \frac{\partial \lambda_1(t)}{\partial \lambda_n(0)} \\
\vdots & \ddots & \vdots \\ \frac{\partial \lambda_n(t)}{\partial \lambda_1(0)} & \dots & \frac{\partial \lambda_n(t)}{\partial \lambda_n(0)}\end{bmatrix}_{\lambda^{(k)}(0)}. \nonumber
\end{equation}

The dynamics of the \textit{influence function matrices} can be derived from the state and costate equation $\dot{x}(t) = H_\lambda^T, \dot{\lambda}(t) = -H_x^T$.
Taking the partial derivatives of these equations w.r.t the initial value of the costate yields
\begin{align}
\frac{\partial}{\partial\lambda(0)} \left[\dot{x}(t) \right] &= \frac{\partial}{\partial\lambda(0)} \left[H_\lambda^T \right] \nonumber\\
\frac{\partial}{\partial\lambda(0)}\left[\dot{\lambda}(t)\right] &= \frac{\partial}{\partial\lambda(0)}\left[-H_x^T\right]. \label{eq:infMat}
\end{align}
With the assumption that $\frac{\partial}{\partial\lambda(0)}[\dot{x}(t)]$, $\frac{\partial}{\partial\lambda(0)}[\dot{\lambda}(t)]$ are continuous and by using the chain rule on right had side of (\ref{eq:infMat}), the differential equations of the \textit{influence function matrices} are obtained as
\begin{equation}
\begin{bmatrix}\dot{P_x}(\lambda^{(k)}(0),t)\\ \dot{P_{\lambda}}(\lambda^{(k)}(0),t) \end{bmatrix} = \begin{bmatrix}\frac{\partial^2 H}{\partial x \partial \lambda} & \frac{\partial^2 H}{\partial \lambda^2}\\ -\frac{\partial^2 H}{\partial x^2} & -\frac{\partial^2 H}{\partial \lambda \partial x} \end{bmatrix}\begin{bmatrix}P_x(t)\\ P_{\lambda}(t)\end{bmatrix},
\label{eq:ve1}
\end{equation}
where $P_x(\lambda^{(k)}(0),0) = 0, P_{\lambda}(\lambda^{(k)}(0),0) = I$.  

On the other side, since the final time of the problem is free, the Hamilitonian should vanish:
\begin{equation}
H(x(\tau),\lambda(\tau),\tau) = 0. \nonumber
\end{equation}
The update rule for the initial costate value and the final time can then be obtained as:
\begin{align}
&\begin{bmatrix}\lambda^{(k+1)}(0)\\ \tau^{(k+1)} \end{bmatrix} = \begin{bmatrix}\lambda^{(k)}(0)\\ \tau^{(k)} \end{bmatrix} \nonumber\\
&-\begin{bmatrix}P_x(\lambda^{(k)}(0),\tau^{(k)}) & \frac{\partial x^{(k)}(\tau^{(k)})}{\partial\tau} \\ \frac{\partial H^{(k)}(\tau^{(k)})}{\partial\lambda^{(k)}(0)} & \frac{\partial H^{(k)}(\tau^{(k)})}{\partial \tau} \end{bmatrix}^{-1}\begin{bmatrix}x^{(k)}(\tau^{(k)})-x_1\\ H^{(k)}(\tau^{(k)})\end{bmatrix},
\label{eq:ve2}
\end{align}
where entries of the matrix on the right hand side can be obtained from:
\begin{align}
\frac{\partial H(\tau^{(k)})}{\partial\lambda_0} &= \frac{\partial H}{\partial x} \frac{dx(\tau^{(k)})}{d\lambda(0)} + \frac{\partial H}{\partial \lambda} \frac{d\lambda(\tau^{(k)})}{d\lambda(0)} \nonumber\\
&= -\dot{\lambda}(\tau^{(k)})^TP_{\lambda}(\tau^{(k)})+\dot{x}(\tau^{(k)})^TP_x(\tau^{(k)}),
\label{eq:ve3}
\end{align}

\begin{equation}
\frac{\partial x^{(k)}(\tau^{(k)})}{\partial \tau^{(k)}} = f(x^{(k)}(\tau^{(k)}), h(x^{(k)}(\tau^{(k)}),\lambda^{(k)}(\tau^{(k)}))),
\label{eq:ve4}
\end{equation}

\begin{eqnarray}
\frac{\partial H(\tau^{(k)})}{\partial\tau^{(k)}} &=& \frac{\partial H}{\partial x} \frac{dx(\tau^{(k)})}{d\tau^{(k)}} + \frac{\partial H}{\partial \lambda} \frac{d\lambda(\tau^{(k)})}{d\tau^{(k)}} \nonumber\\
&=& 0.
\label{eq:ve5}
\end{eqnarray}

Finally, the TPBVP solver with the variation of extremals based method can be summarized as Algorithm \ref{alg:ve}.
\begin{algorithm}
\caption{$\pi\gets$VE-based TPBVPsolver$(x_0,x_1)$}\label{alg:ve}
\begin{algorithmic}[1]
\State $\textmd{Initialize}~\lambda^{(0)}(0),~\tau^{(0)}~\textmd{and set}~k=0$ \Comment{Section \ref{subsec:OC}}
\Repeat
\State $k = k+1;$
\State Integrate $x^{(k)},\lambda^{(k)},P_x,P_{\lambda}$ \Comment{(\ref{eq:TPBVP2}), Equation (\ref{eq:ve1})}
\State Calculate $\lambda^{(k+1)}(0),\tau^{(k+1)}$\Comment{Equation (\ref{eq:ve2})-(\ref{eq:ve5})}
\Until{Converge}
\State Calculate $u(t)$ \Comment{Equation (\ref{eq:con})}
\State \textbf{return} $\pi \gets (x^{(k)}(\cdot),u(\cdot), \tau^{(k)})$
\end{algorithmic}
\end{algorithm}

\subsection{Discussions}
Both methods presented in this section find the solution to (\ref{eq:TPBVP2}) iteratively. At each iteration, the SA-based method computes the optimal solution to an linear approximation of the original problem linearized at the solution of the previous iteration, while the VE-based method updates the estimate of the initial costate and the final time. It should be pointed out that these solvers guarantee local (not global) optimality, because the underlying TPBVP is derived as a necessary (not sufficient) condition for optimality. However, this locality would not make a significant impact on the overall motion planning, since the TPBVP itself is posed for optimally linking a small segment of the overall plan.

Compared to an existing approach that treated \textit{linearized} dynamics~\cite{Dustin13} and thus did not require iterative process for solving a TPBVP,  the proposed methods need to solve additional $k_{iter}$ first-order ODEs, where $k_{iter}$ is the number of iterations for convergence, in computing a single optimal trajectory segment in the RRT* framework. However, the proposed schemes can take into account nonlinear kinodynamic constraints, which was not accurately realized in the previous work.

Note also that the two methods exhibit different characteristics from the computational point of view.
First, the storage requirement of the VE-based method has an advantage over that of the SA-based method: at every iteration, the VE solver only needs to store 2 $(n \times n)$ matrices, $P_x(t)$ and $P_{\lambda}(t)$, while the SA solver needs to store the trajectories, $\lambda_p^{(k)}(t)$, $x^{(k)}(t)$ and $\lambda^{(k)}(t)$ for $t\in[0,\tau]$ and a $(n \times n)$ matrix, $G(t)$.
On the other side, the SA solver demands integration of $4n$ first-order ODEs in  (\ref{eq:lp}), (\ref{eq:sa1}) and (\ref{eq:alg2}) at each iteration ($n^2$ first-order ODEs in (\ref{eq:sa2}) needs to be integrated only at first iteration and then reused.), while the VE solver needs to integrate $2n(n+1)$ first-order ODEs in (\ref{eq:ve1}) and (\ref{eq:TPBVP2}) in each iteration.
Thus, assuming both methods converge with similar number of iterations, the VE-based method exhibits better memory complexity, while the SA-based one gives better time complexity, in general. However, detailed convergence characteristics such as convergence time might be problem dependent.

\section{Numerical Examples} \label{sec:numerics}
This section demonstrates the validity of the proposed optimal kinodynamic planning methods on three numerical examples:
The first pendulum swing-up example compares the proposed scheme with a state-of-the-art method based on linearized dynamics.
The second example on a two-wheeled mobile robot in a cluttered environment demonstrates the asymptotic optimality of the algorithms.
The third example on a SCARA type robot arm investigates the characteristics of the optimized plans with varying cost functionals.

\subsection{Pendulum Swing-Up} \label{sec:pendulum}

\begin{figure}[t]
\centering
\subfigure[SA-based solver, $R=1$]{
\includegraphics*[width=4.2cm,viewport=35 20 390 300]{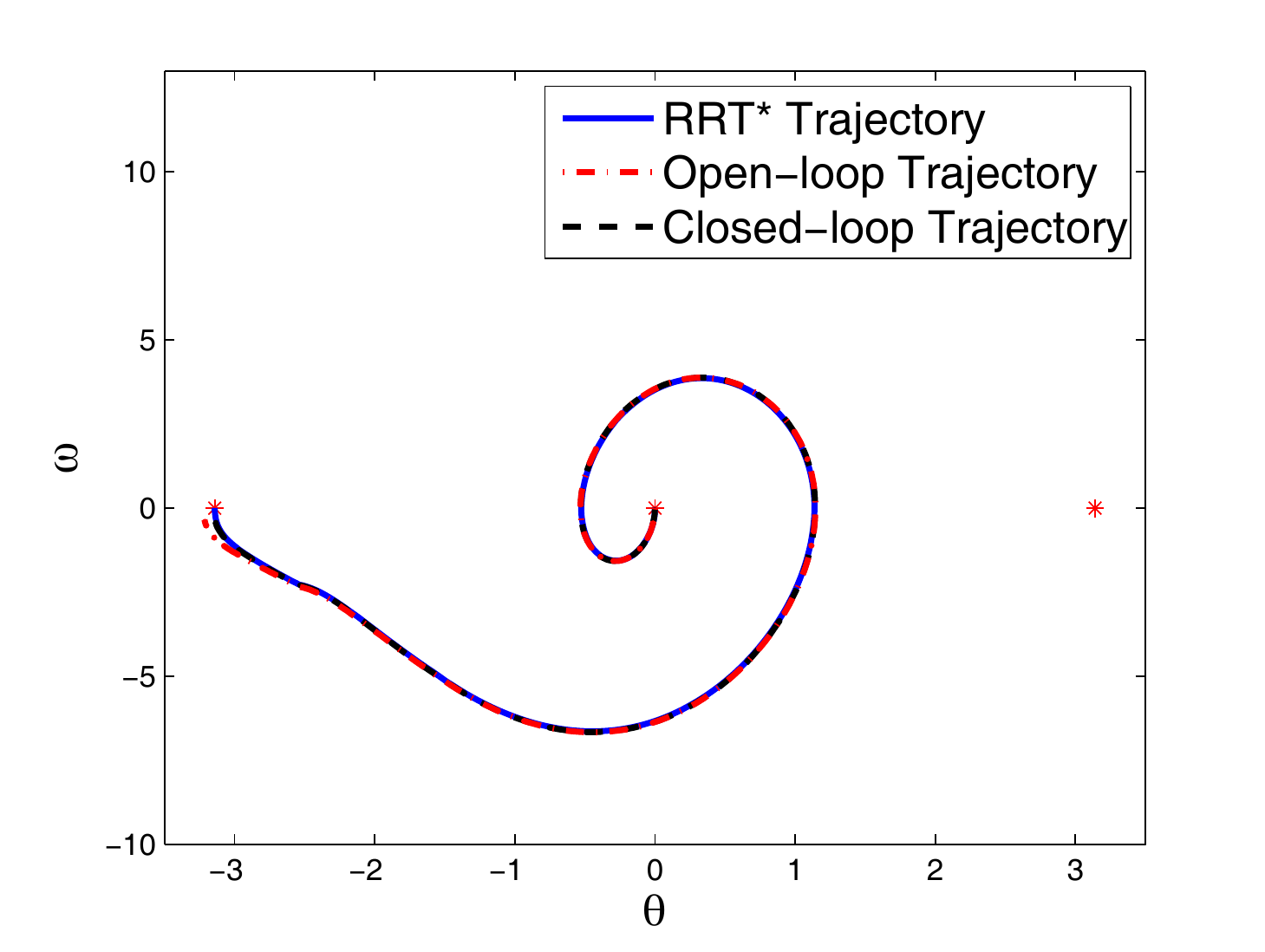}}
\subfigure[Solver for linearized dynamics, $R=1$]{
\includegraphics*[width=4.2cm,viewport=35 20 390 300]{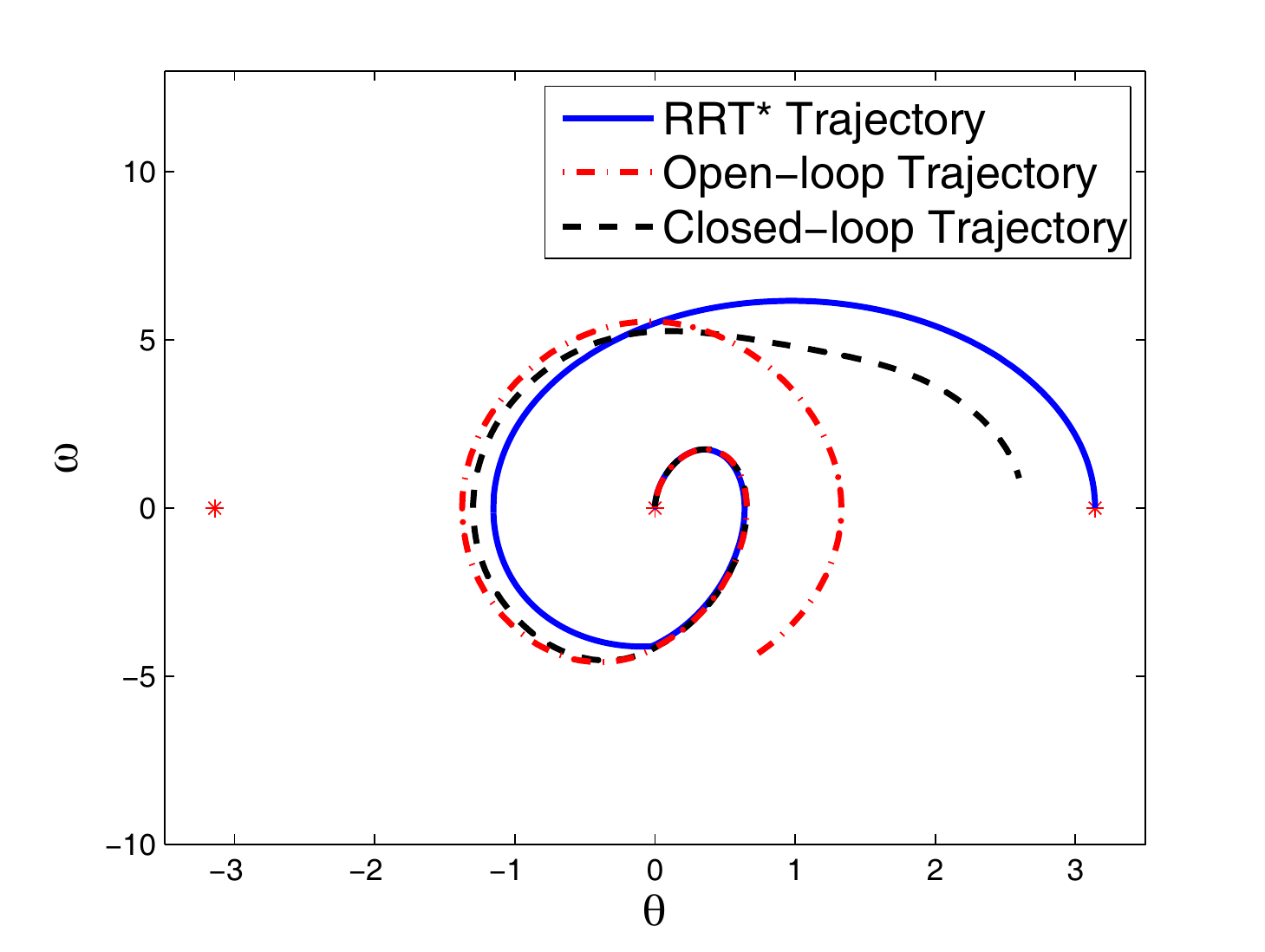}}
\subfigure[SA-based solver, $R=10$]{
\includegraphics*[width=4.2cm,viewport=35 20 390 300]{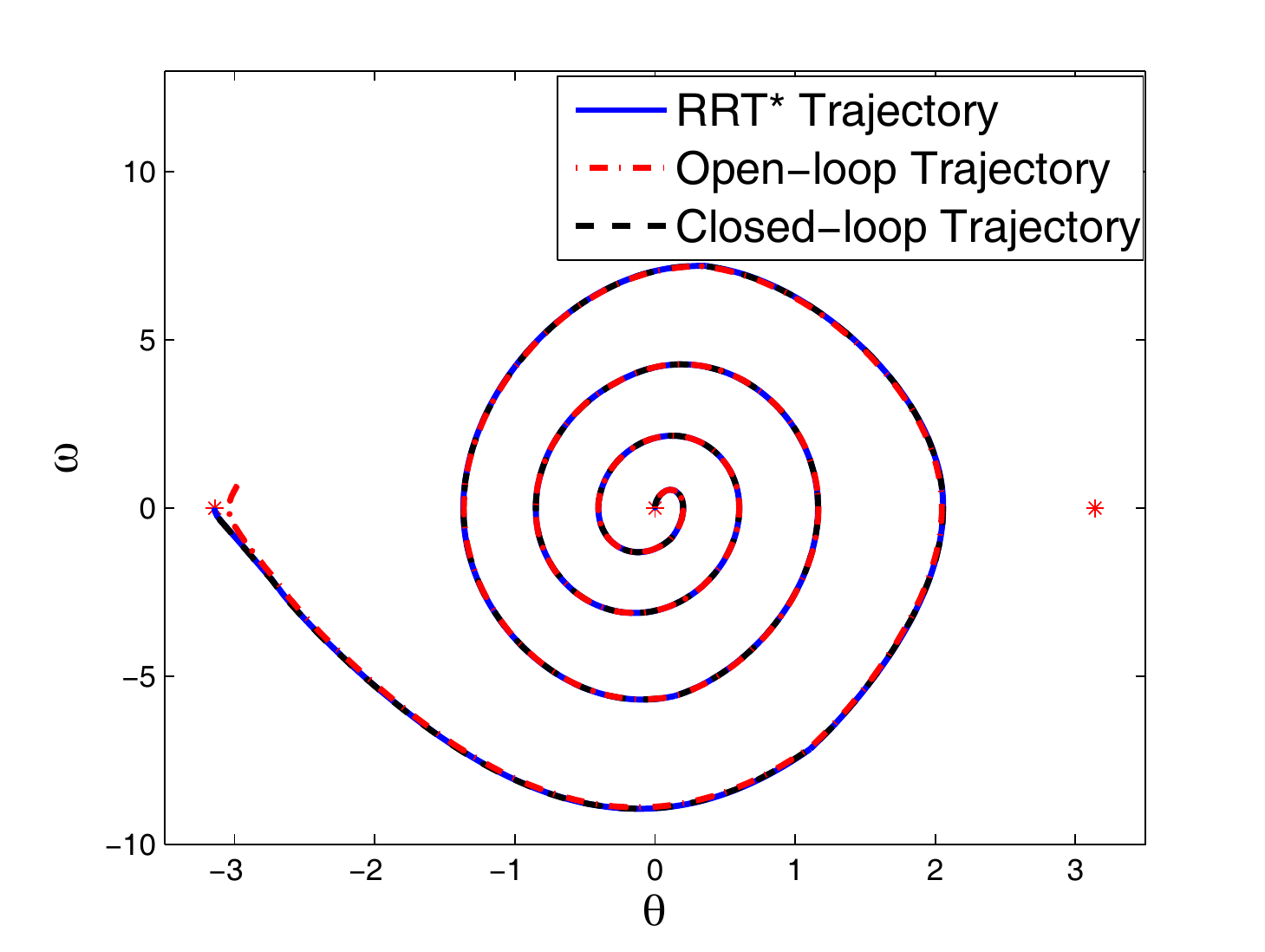}}
\subfigure[Solver for linearized dynamics, $R=10$]{
\includegraphics*[width=4.2cm,viewport=35 20 390 300]{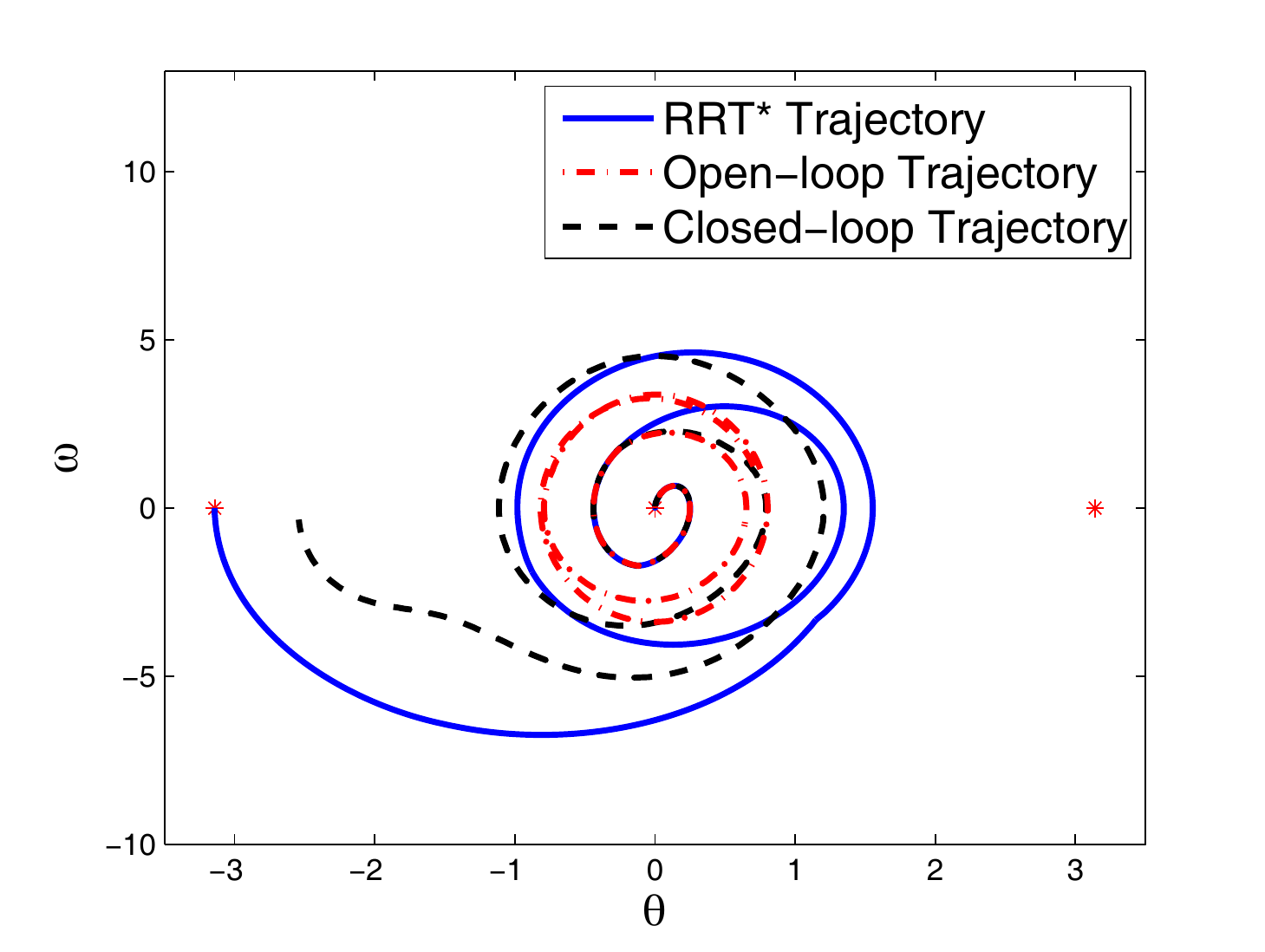}}
\caption{State trajectories for different TPBVP solvers with varying $R$ for pendulum swing up example.}
\label{result2}
\end{figure}

\begin{figure}[t]
\centering
\subfigure[SA-based solver, $R=1$]{
\includegraphics*[width=4.2cm,viewport=35 20 390 300]{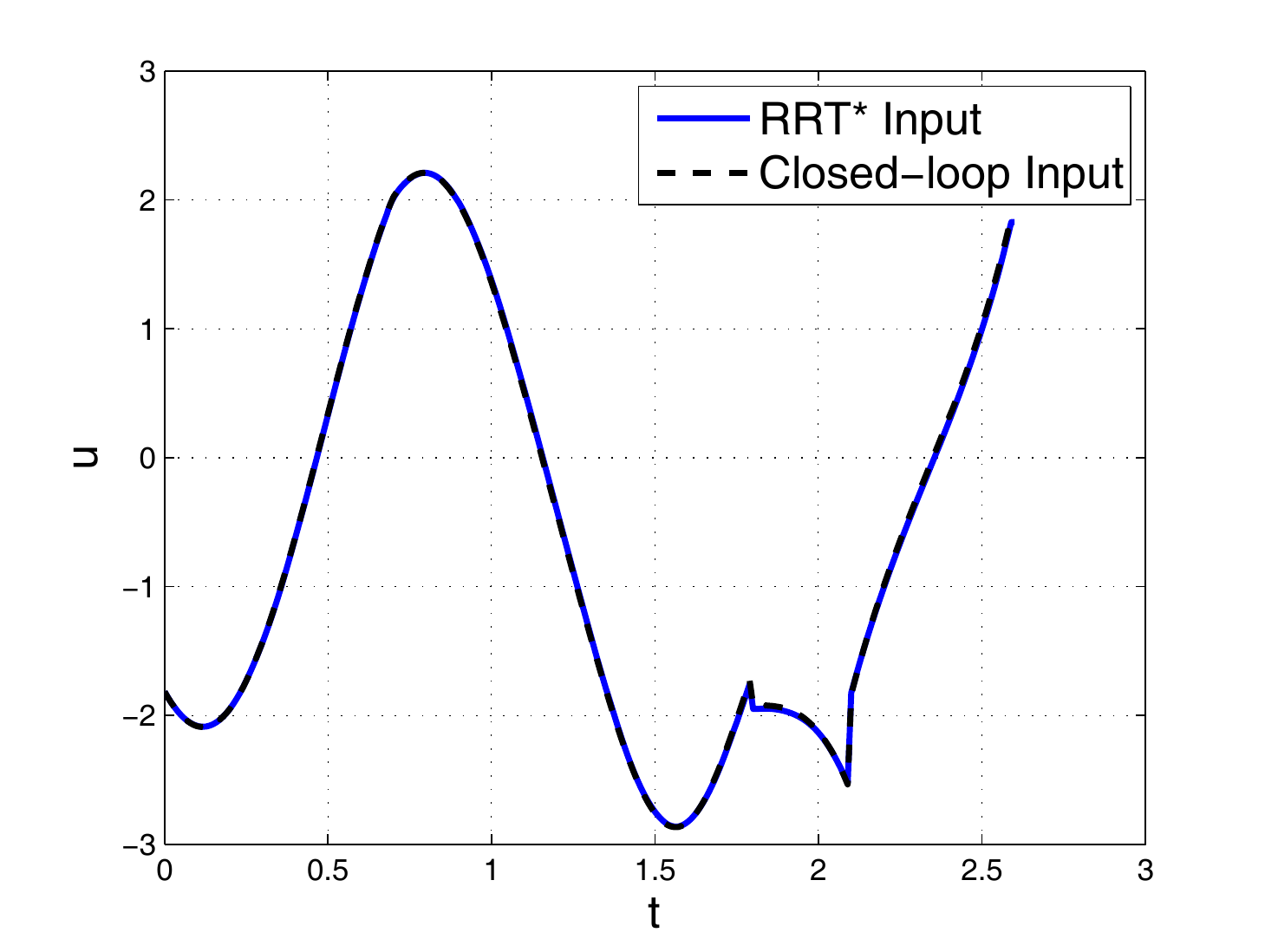}}
\subfigure[Solver for linearized dynamics, $R=1$]{
\includegraphics*[width=4.2cm,viewport=35 20 390 300]{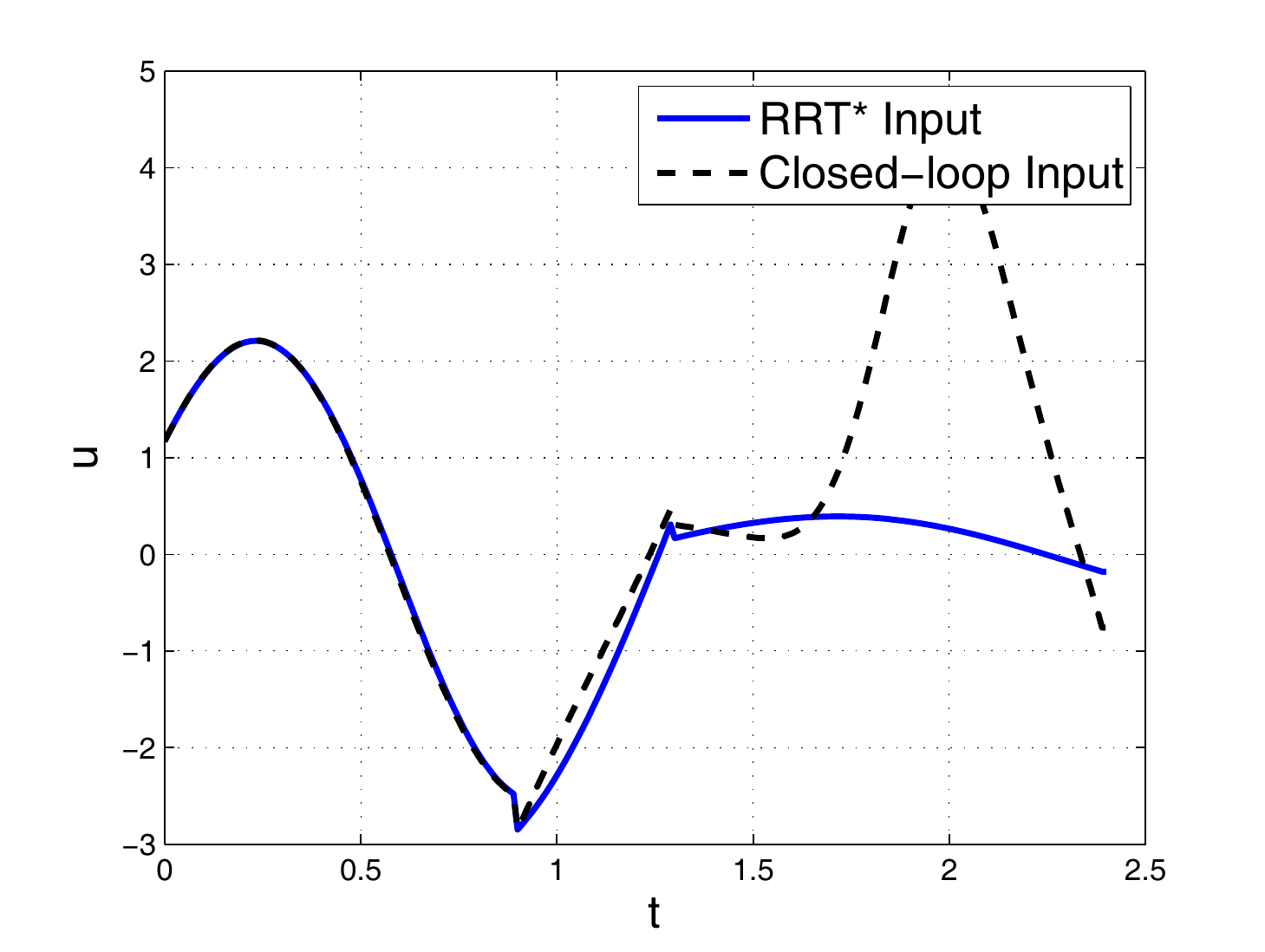}}
\subfigure[SA-based solver, $R=10$]{
\includegraphics*[width=4.2cm,viewport=35 20 390 300]{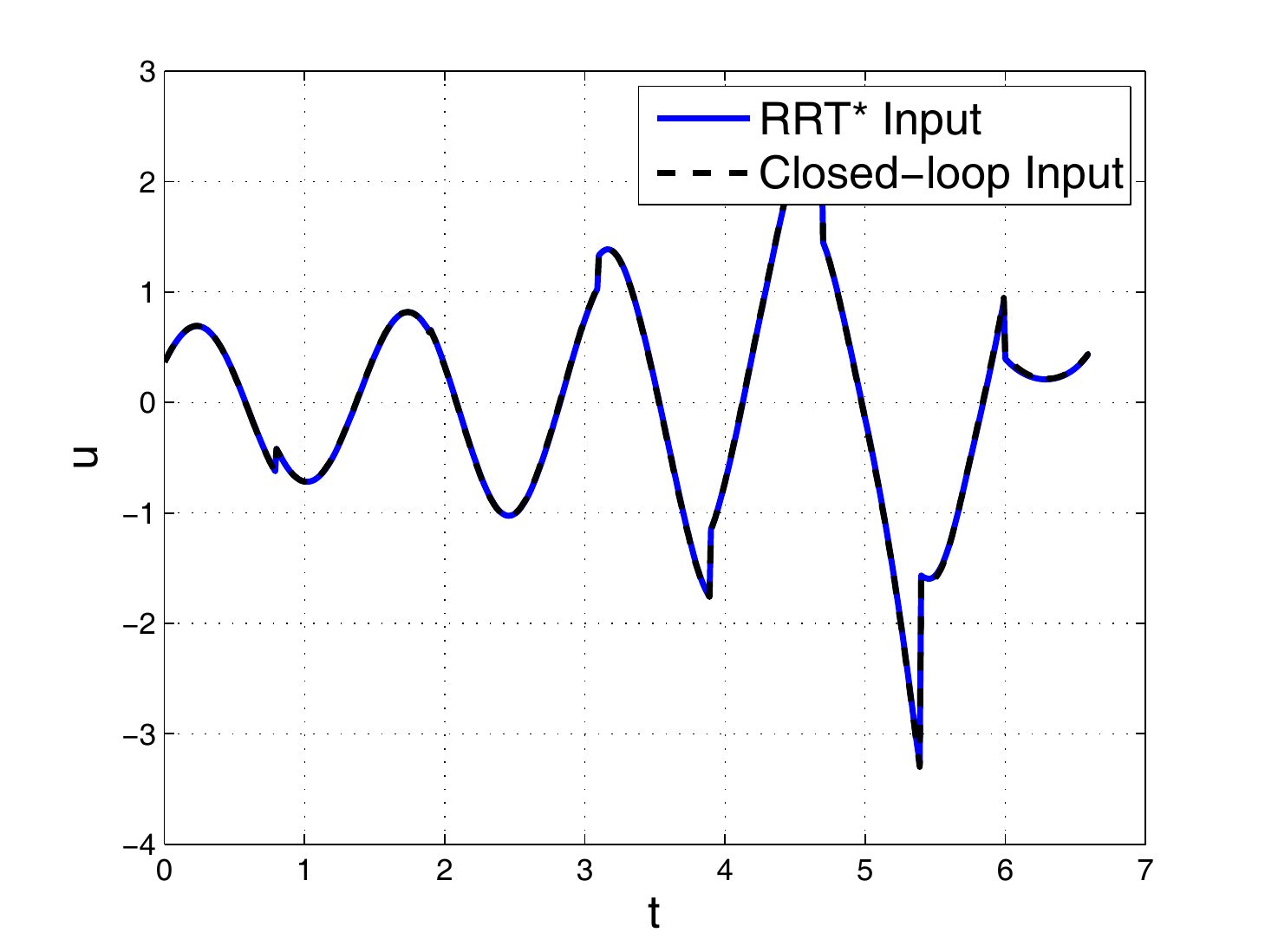}}
\subfigure[Solver for linearized dynamics, $R=10$]{
\includegraphics*[width=4.2cm,viewport=35 20 390 300]{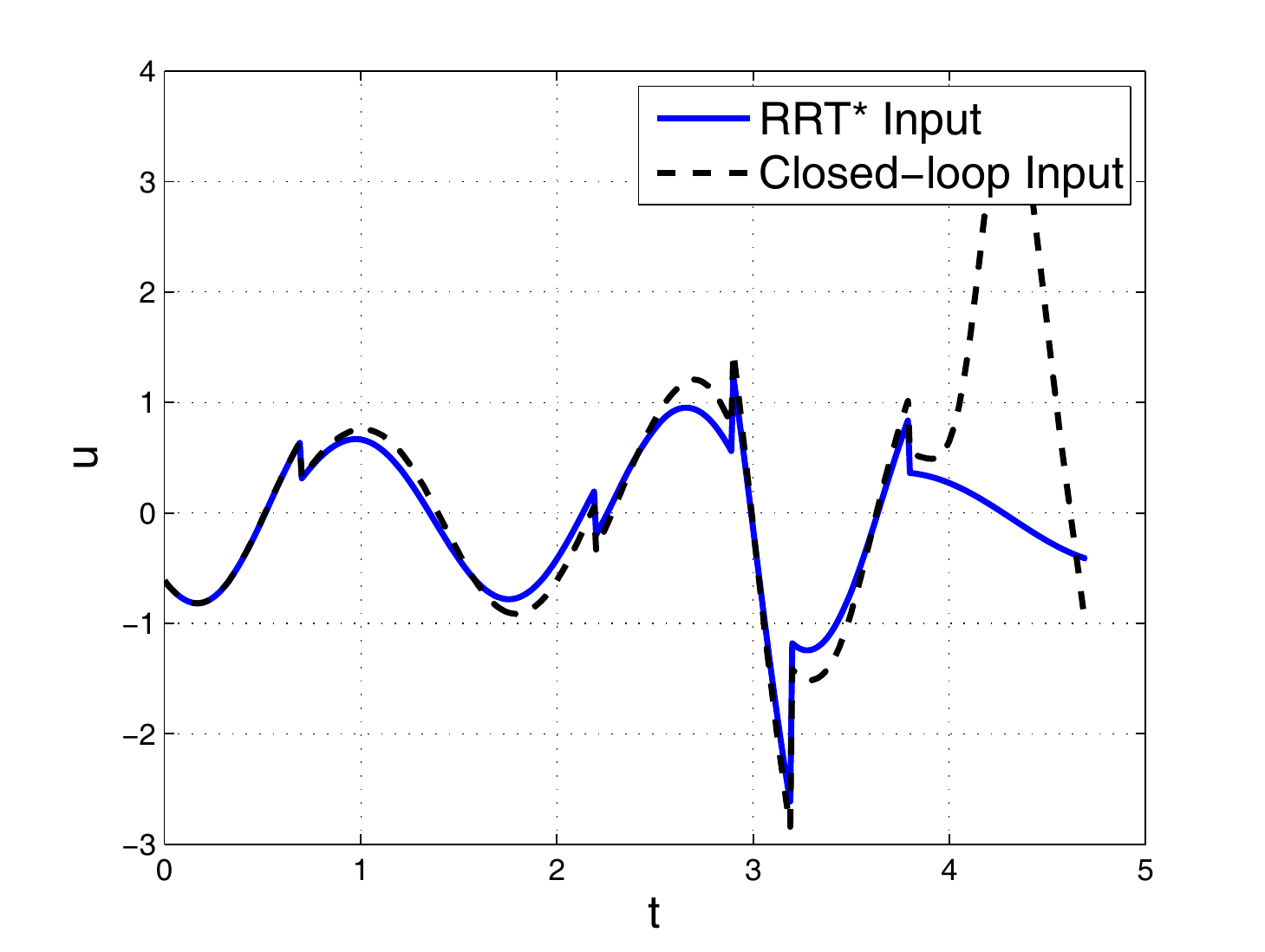}}
\caption{Planned control input vs. Executed(feedforward+feedback) control input containing feedback for different TPBVP solvers with varying $R$ for pendulum swing up example.}
\label{result4}
\end{figure}

The control objective of the pendulum swing-up problem is to put the pendulum in an upright position from its downward stable equilibrium.
The dynamics of the system is given as
\begin{equation}
I\ddot{\theta} + b\dot{\theta} + mgl_c\sin\theta = u.
\end{equation}
The angular position and velocity, $x(t) = [\theta(t)~\dot{\theta}(t)]^T$ are the state variables and the torque, $u(t)$ is the control input to the system. The cost of the trajectory is given by  (\ref{eq:cost}).
The initial and goal states are given as $x_{init}=[0~0]^T, x_{goal}=\{[\pi~0]^T,~[-\pi~0]^T\}$.
To comparatively investigate the capability of RRT* variants in handling nonlinearity in dynamics, the proposed iterative methods are compared with kinodynamic-RRT*  \cite{Dustin13} that takes into account the first-order Taylor approximation of the dynamics.

Fig. \ref{result2} illustrates state trajectories in the phase plane; an open-loop and a closed-loop trajectories are depicted as well as the planned trajectory. The open-loop trajectory is generated by using the planned control input trajectory $u(t)$ as a feedfoward control term, while the closed-loop trajectory is produced by adding an LQR trajectory stabilizer ~\cite{tedrake2009underactuated} that utilizes feedback control in the form of $u(t) - K \bar{x}$ with LQR gain $K$ where $\bar{x}$ being the deviation from the planned trajectory. It was found that both of the proposed iterative solvers produce very similar results; thus only the result of the SA-based solver is presented herein.

The blue-solid line represents the planned trajectories from the algorithm using the SA-based solver (left) and the solver for linearized dynamics (right);
the red-dashed-dot and black-dashed lines denote the open-loop and closed-loop state trajectories, respectively.
Note that as the linearization is only valid around the nodes of the RRT* tree, the optimized trajectory segment with using the linearization-based scheme~\cite{Dustin13} is dynamically infeasible for the original (nonlinear) system as the state moves far away from the nodes.
As shown in Fig.\ref{result2} (b) and (d), the system may not follow the planned trajectory with open-loop control and even with a feedback trajectory stabilizer; while, as shown in Fig.\ref{result2} (a) and (c), the planned trajectories from the algorithm using the iterative TPBVP solver is followed not only by the feedback control but also by the open-loop control.

The control input histories are shown in Fig. \ref{result4}.
The blue-solid line represents the input from the planning algorithm and the black-dashed line denotes the input with feedback control implementation.
In case the planned trajectory is inconsistent with the real dynamics, a large amount of control effort is required in the feedback controller, which results in cost increase.
This can be observed in the case where the linearization-based planner is used; note that the control effort for the proposed method is significantly smaller than that of the linearization-based method.
Also, Table \ref{result_tab1} represents average costs of ten simulations for varying cost functions with feedback controller;
the `planned cost' means the cost returned from the algorithm, and the `executed cost' represents the cost which include control effort from the trajectory stabilization feedback controller.
It is shown that the cost achieved by using proposed method is smaller and more consistent with planned cost.

\begin{table}[t]
\caption{Average costs from ten simulations}
\label{result_tab1}
\centering
\begin{tabular} {c|cc|cccccccc}
\hline\hline
 &\multicolumn{2}{c}{SA-based solver}&\multicolumn{2}{c}{Solver for linearized dynamics}\\
\cline{2-3} \cline{4-5}
 & planned cost & executed cost & planned cost & executed cost \\
\hline
 R=1 & 6.4621 & 6.4727 & 4.5122 & 8.4837 \\
 R=5 & 18.9790 & 18.9929 & 9.7229 & 31.2797 \\
 R=10 & 33.0323 & 33.7034 & 12.9783 & 55.9197 \\
 R=15 & 52.8430 & 52.9865 & 15.1409 & 83.6147 \\
\hline\hline
\end{tabular}
\end{table}

\subsection{Two-Wheeled Mobile Robot} \label{sec:TWMR}
The second example addresses design of the trajectory of a two-wheeled mobile robot.
The states and inputs of the system are $x = [p_x~p_y~\theta~v~w]^T$ and $u = [F_1~F_2]^T$, where $p_x,p_y$ and $\theta$ represent robot position and orientation, respectively, $v,~w$ denote the linear and angular velocity of the robot, and $F_1,~F_2$ are the force from each wheel. The dynamic equation of the system is given as
\begin{equation}
\begin{bmatrix}\dot{p}_x\\ \dot{p}_y\\ \dot{\theta}\\ \dot{v}\\ \dot{w}\end{bmatrix} = \begin{bmatrix}v\cos\theta\\ v\sin\theta\\ w\\ F_1+F_2\\ F_1-F_2\end{bmatrix} = \begin{bmatrix}v\cos\theta\\ v\sin\theta\\ w\\ u_1+u_2\\ u_1-u_2\end{bmatrix}.\nonumber
\end{equation}
The initial and the goal states are given as $x(0) = [0.5~0.5~\pi/4~1~0]^T$ and $x(\tau) = \{[p_x~p_y~\theta~v~w]^T|23\leq p_x\leq 24,~9\leq p_y\leq 10,~0\leq \theta\leq \pi/2,~0.8\leq v\leq 1.2,~-0.2\leq w\leq 0.2\}$, respectively, with free final time $\tau$.
The cost functional is given in (\ref{eq:cost}) with $R = 20\text{diag}(1,1)$.

\begin{figure}[t]
\vspace*{2mm}
\centering
\subfigure[200 nodes]{
\includegraphics*[width=8cm,viewport=54 92 395 235]{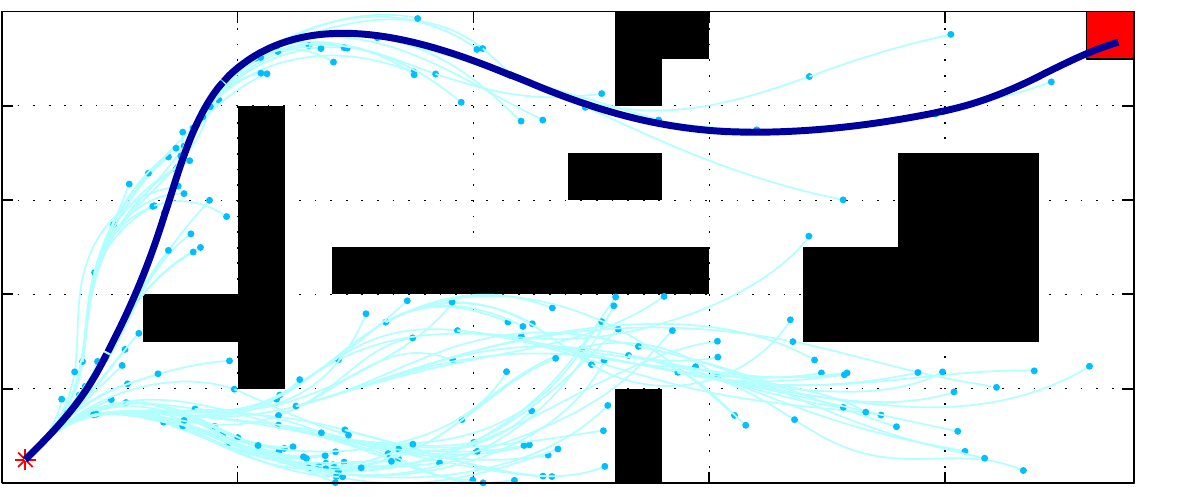}}
\subfigure[500 nodes]{
\includegraphics*[width=8cm,viewport=54 92 395 235]{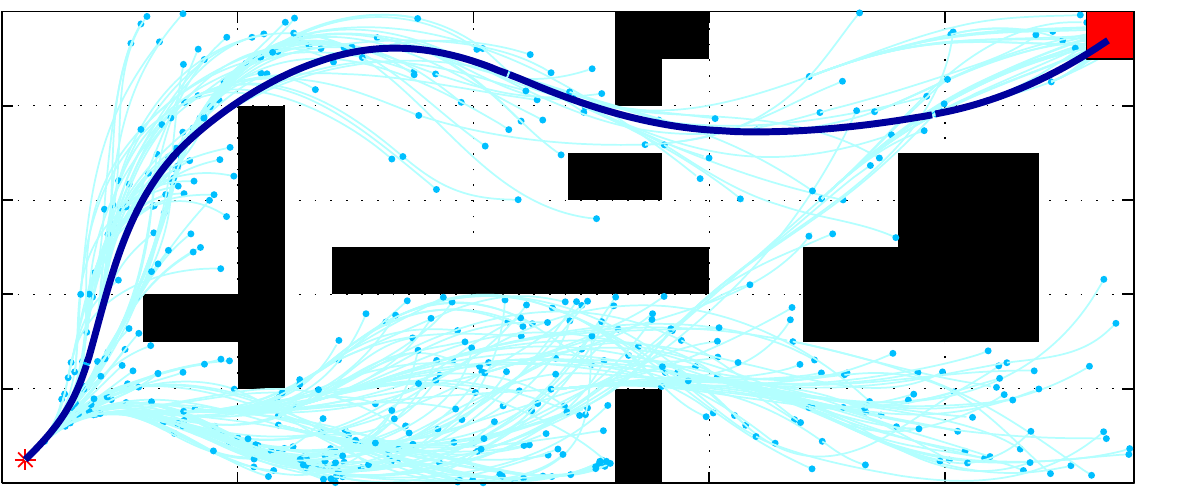}}
\subfigure[1000 nodes]{
\includegraphics*[width=8cm,viewport=54 92 395 235]{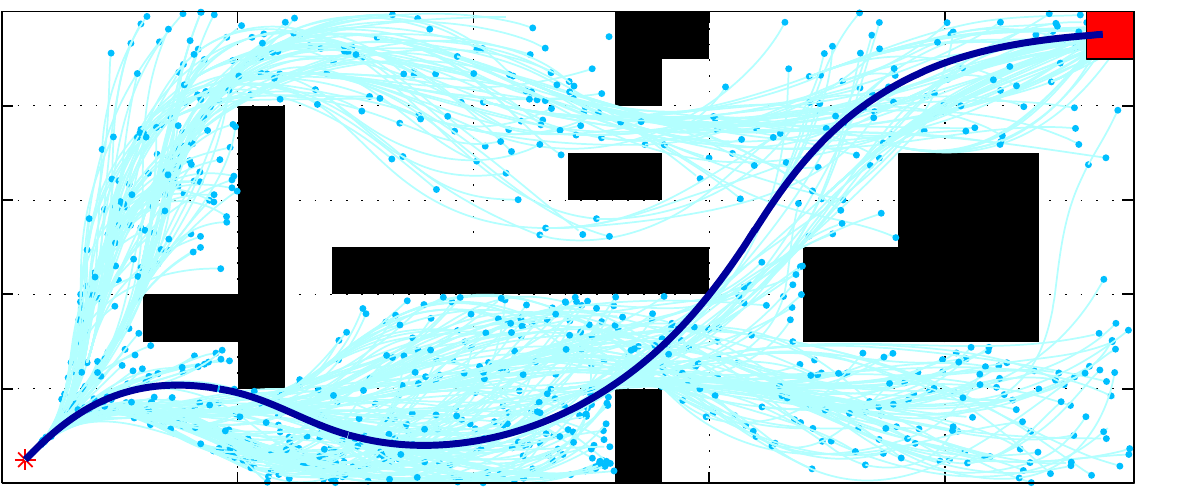}}
\subfigure[5000 nodes]{
\includegraphics*[width=8cm,viewport=54 92 395 235]{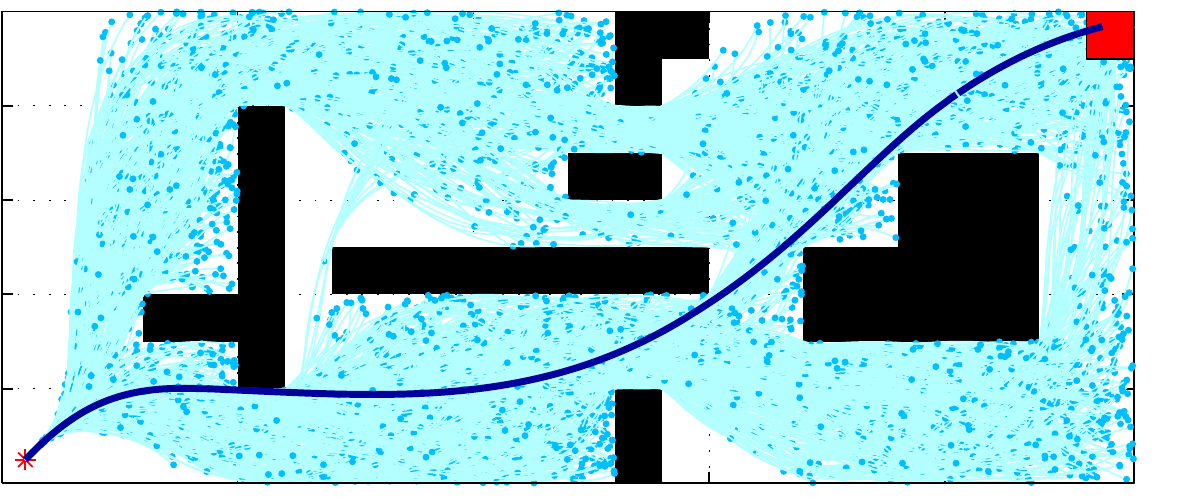}}
\caption{Two-dimensional representation of evolution of RRT* tree for mobile robot example: when the number of nodes are (a) 200, (b) 500, (c) 1000 and (d) 5000 for two-wheeled mobile robot example.}
\label{fig:result1}
\end{figure}

\begin{table}[t]
\caption{Cost of the best trajectories in the tree out of 100 trials.
}
\label{result_tab2}
\centering
\begin{tabular} {ccccccccccc}
\hline\hline

\# of nodes  & 300 & 500 & 1000 & 3000 & 5000 \\
\hline
 \# of feasible  & 81 & 100 & 100 & 100 & 100 \\
 Mean  & inf & 21.8125 & 20.5124 & 19.5235 & 19.1839 \\
 Variance  & NaN & 2.2570 & 0.7864 & 0.2261 & 0.1687 \\
\hline\hline
\end{tabular}
\end{table}

\begin{figure}[t]
\vspace*{2mm}
\centering
\subfigure[Proposed solver]{
\includegraphics*[width=8cm,viewport=54 92 395 235]{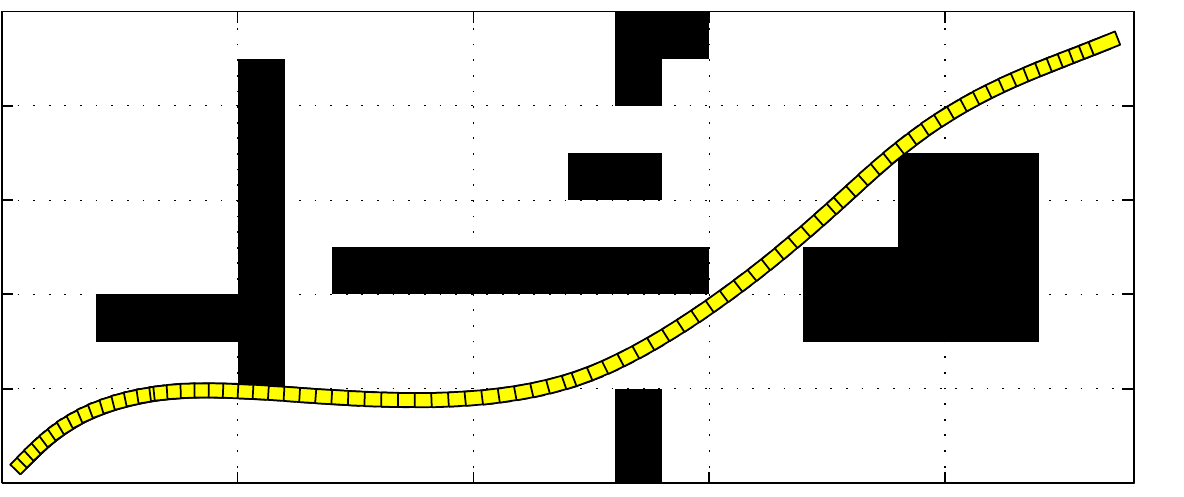}}
\subfigure[Solver for linearized dynamics]{
\includegraphics*[width=8cm,viewport=54 92 395 235]{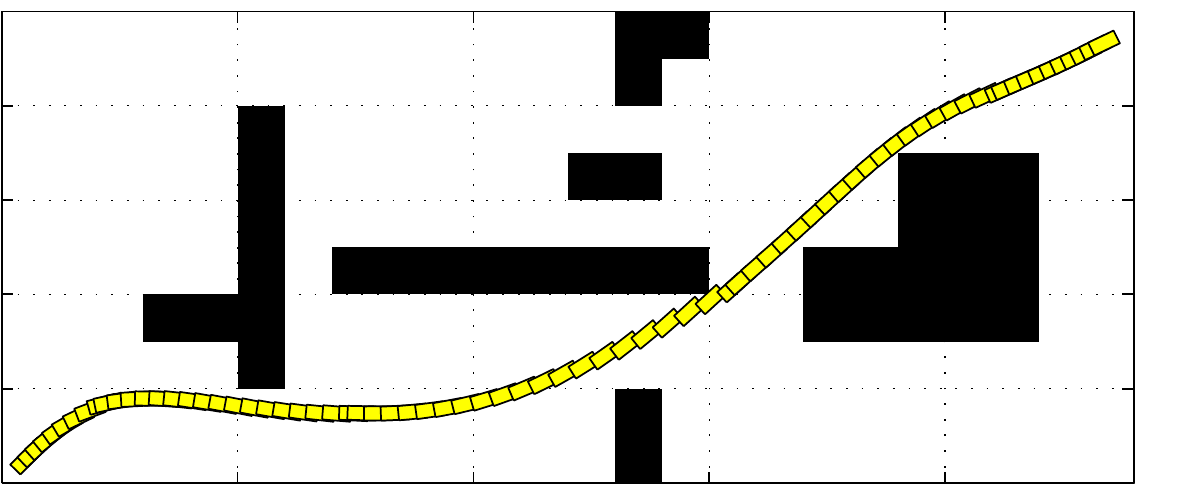}}
\caption{Resulting trajectories with proposed solver and solver for linearized dynamics for two-wheeled mobile robot example.}
\label{fig:result2}
\end{figure}

As the two proposed iterative methods have also shown similar performances, only the result for the VE-based solver case is depicted in this example.
Fig. \ref{fig:result1} represents the progression of the RRT* tree, by projecting the tree in five-dimensional space onto the two-dimensional position space, $(p_x,p_y)\in \mathbb{R}^2$.
The red star and the square represent the initial position and the goal region; the thick dark-blue line represents the best trajectory found up to the corresponding progression.
Observe that the resulting trajectories connect the initial and the goal state smoothly. 
Also, as the number of nodes in the tree increases, the tree fills up the feasible state space and finds out a lower-cost trajectory.

Table \ref{result_tab2} reports the  result of a Monte-Carlo simulation with 100 trials; it shows the number of trials finding a feasible trajectories, the mean and variance of the minimum costs in each iteration.
It is found that the mean and variance of the cost decreases as the number of nodes increases, implying convergence to the same solution. 

Finally, Fig. \ref{fig:result2} depicts the resulting trajectories with VE-based solver and the same linearization-based solver as the first example \cite{Dustin13}.
It is observed that with the solver for linearized dynamics, there is a portion of sideway-skid moving at the middle of trajectory, which is dynamically infeasible for the robot.
On the contrary, the result of the proposed solver shows feasible moving in the whole trajectory.

\subsection{SCARA type Robot Arm}
\begin{figure}[t]
\centering
\includegraphics[width=7cm]{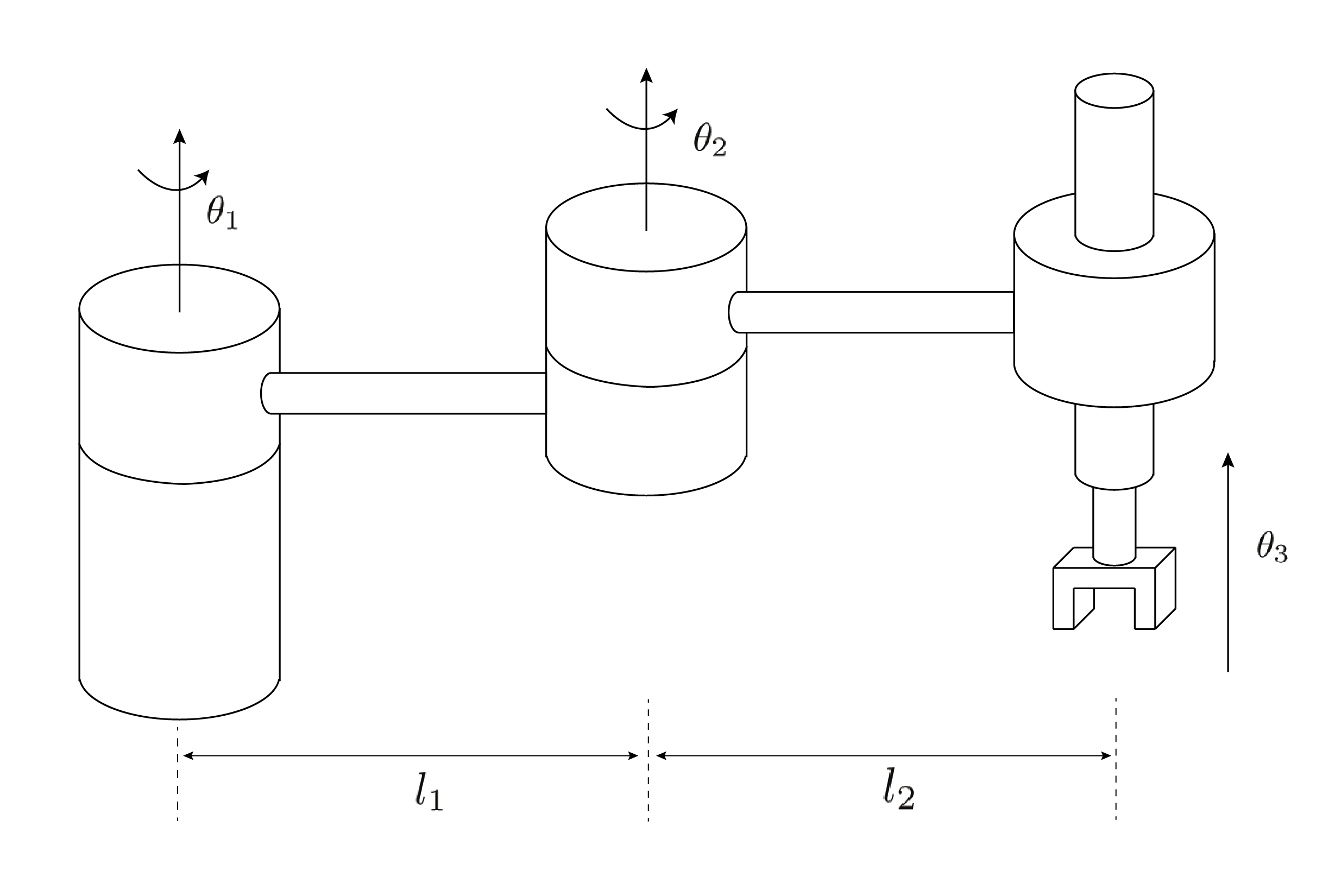}
\caption{SCARA Robot}
\label{fig:scara}
\end{figure}
The third example is about generating a motion plan of a three-degree-of-freedom SCARA robot with two rotational joints (represented by $\theta_1$ and $\theta_2$), and one prismatic joint ($\theta_3$) as shown in Fig. \ref{fig:scara}.
The dynamics of the robot is given as,
\begin{equation}
M(\theta)\ddot{\theta} + C(\theta,\dot{\theta})\dot{\theta} +N(\theta,\dot{\theta}) = u, \nonumber
\end{equation}
where
\begin{align}
M(\theta) = \begin{bmatrix}\alpha+\beta+2\gamma\cos\theta_2 & \beta+\gamma\cos\theta_2 & 0\\ \beta+\gamma\cos\theta_2 & \beta & 0 \\ 0 & 0 & m_3 \end{bmatrix}, \nonumber\\
C(\theta,\dot{\theta}) = \begin{bmatrix}-\gamma\sin\theta_2\dot{\theta}_2 & -\gamma\sin\theta_2(\dot{\theta}_1+\dot{\theta}_2) & 0\\ \gamma\sin\theta_2\dot{\theta}_1 & 0 & 0 \\ 0 & 0 & 0 \end{bmatrix}, \nonumber\\
N(\theta,\dot{\theta}) = \begin{bmatrix} 0 \\ 0 \\ m_3g \end{bmatrix}\nonumber
\end{align}
and
\begin{align}
\alpha = I_{z1}+r_1^2m_1+l_1^2m_2+l_1^2m_3, \nonumber\\
\beta = I_{z2}+I_{z3}+l_2^2m_3+m_2r_2^2, \nonumber\\
\gamma = l_1l_2m_3+l_1m_2r_2.\nonumber
\end{align}
$m_1,~m_2,~m_3$ and $l_1$, $l_2$ represent mass and length of each link, respectively.
Also, $r_1$, $r_2$ denote length between a joint axis and the center of mass of each link and $I_{z1},~I_{z2},~I_{z3}$ represent the moment of inertia about the rotation axis.
The system has a six-dimensional state vector, $x=[\theta_1~\theta_2~\theta_3~\dot{\theta}_1~\dot{\theta}_2~\dot{\theta}_3]^T$, and a three-dimensional control inputs, $u = [\tau_1~\tau_2~f_3]^T$ representing the torques of the two rotational joints and the force of the prismatic joint.

\begin{figure}[t]
\vspace*{2mm}
\centering
\subfigure[$t = 0$ sec.]{
\includegraphics*[width=3.9cm,viewport=85 90 300 285]{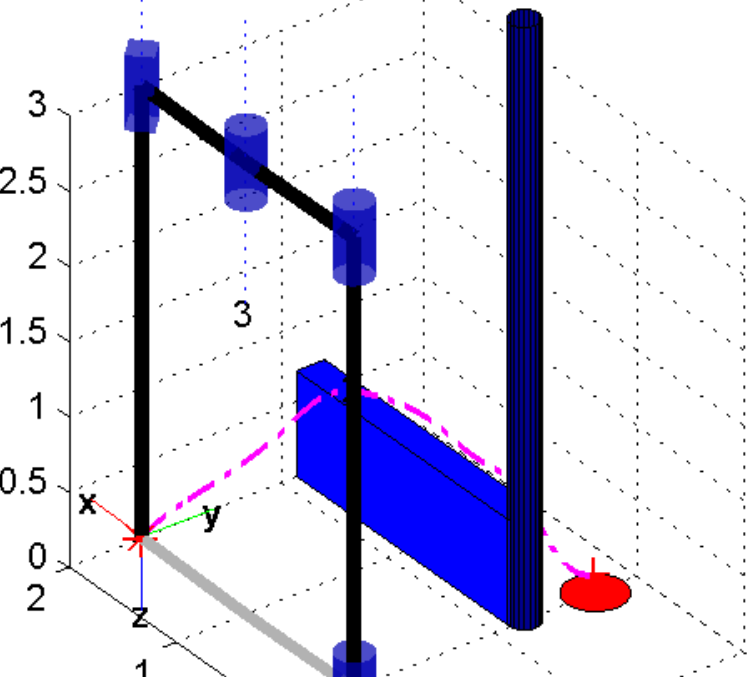}}
\subfigure[$t = 0.7$ sec.]{
\includegraphics*[width=3.9cm,viewport=85 90 300 285]{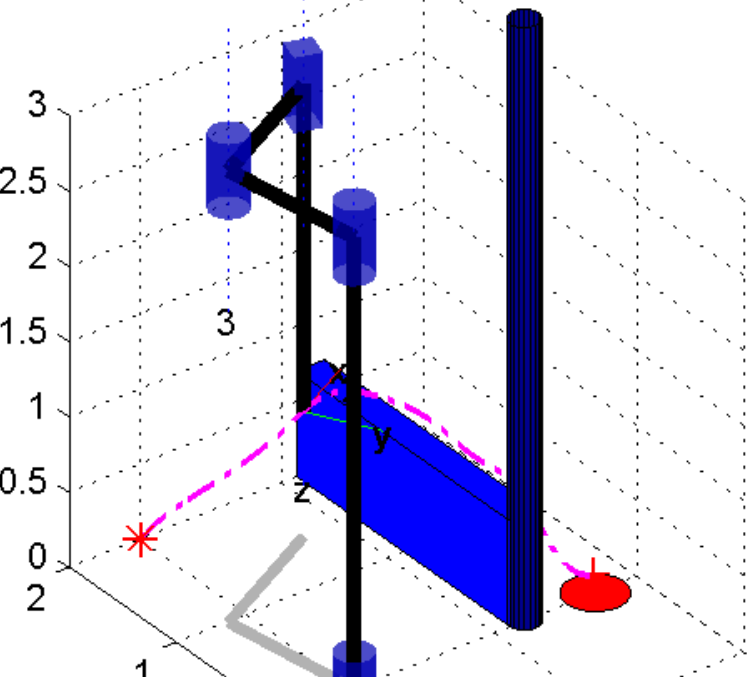}}
\subfigure[$t = 1.45$ sec.]{
\includegraphics*[width=3.9cm,viewport=85 90 300 285]{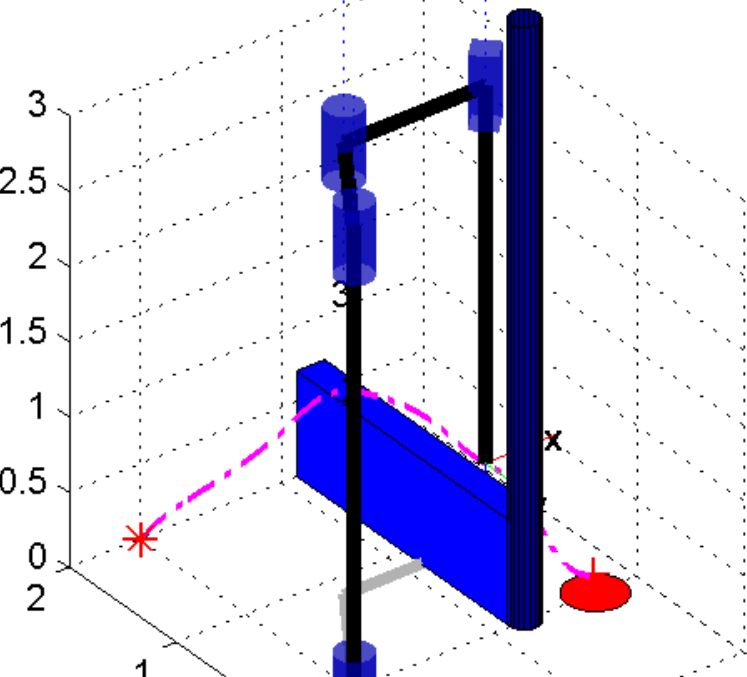}}
\subfigure[$t = 2.15$ sec]{
\includegraphics*[width=3.9cm,viewport=85 90 300 285]{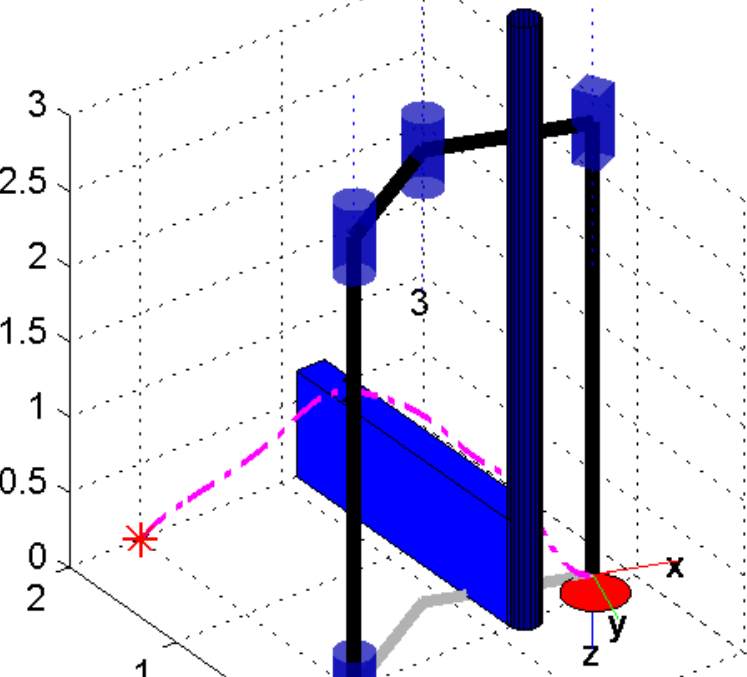}}
\caption{The motion of the SCARA robot corresponding to the resulting state trajectory when 5000 nodes are added to the tree with $R =  0.05\text{diag}(1, 1, 0.5)$.}
\label{fig:result3}
\end{figure}

\begin{figure}[t]
\vspace*{2mm}
\centering
\subfigure[$t = 0$ sec.]{
\includegraphics*[width=3.9cm,viewport=85 90 300 285]{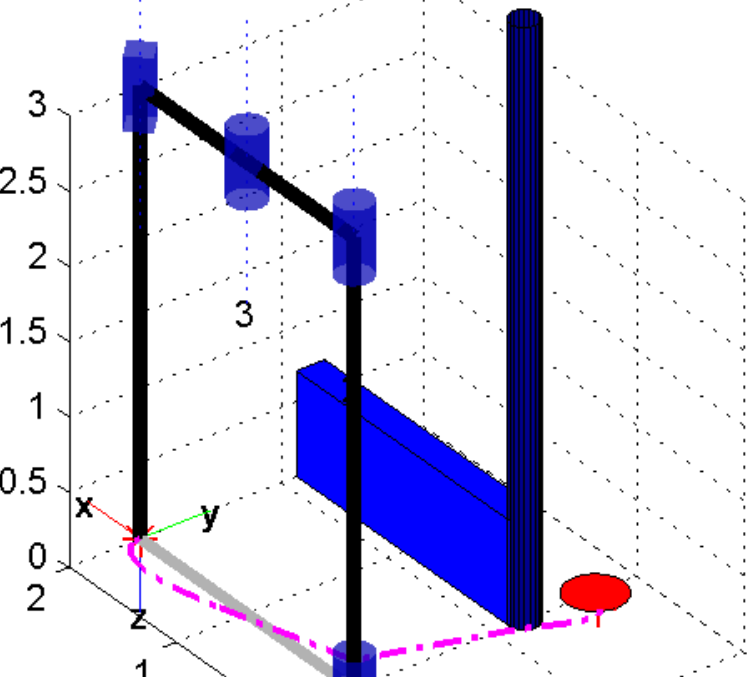}}
\subfigure[$t = 0.4$ sec.]{
\includegraphics*[width=3.9cm,viewport=85 90 300 285]{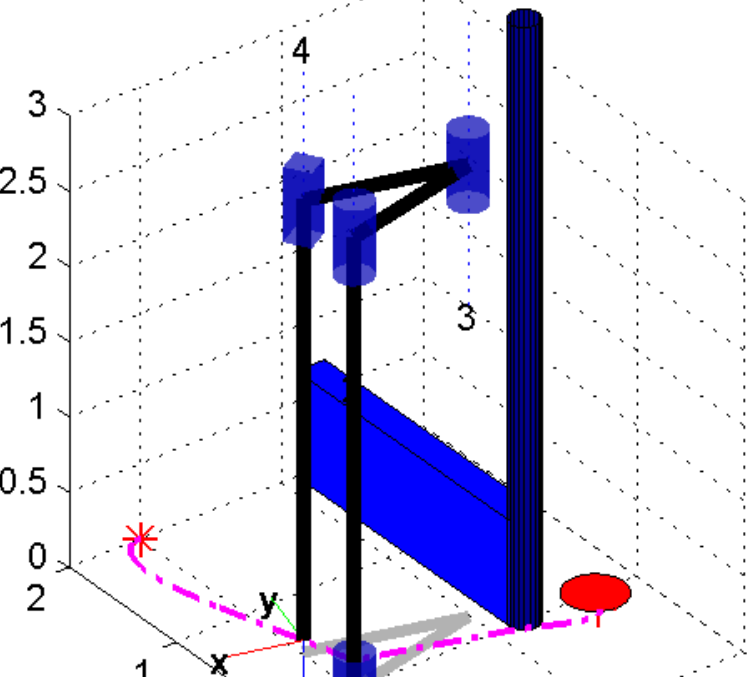}}
\subfigure[$t = 0.85$ sec.]{
\includegraphics*[width=3.9cm,viewport=85 90 300 285]{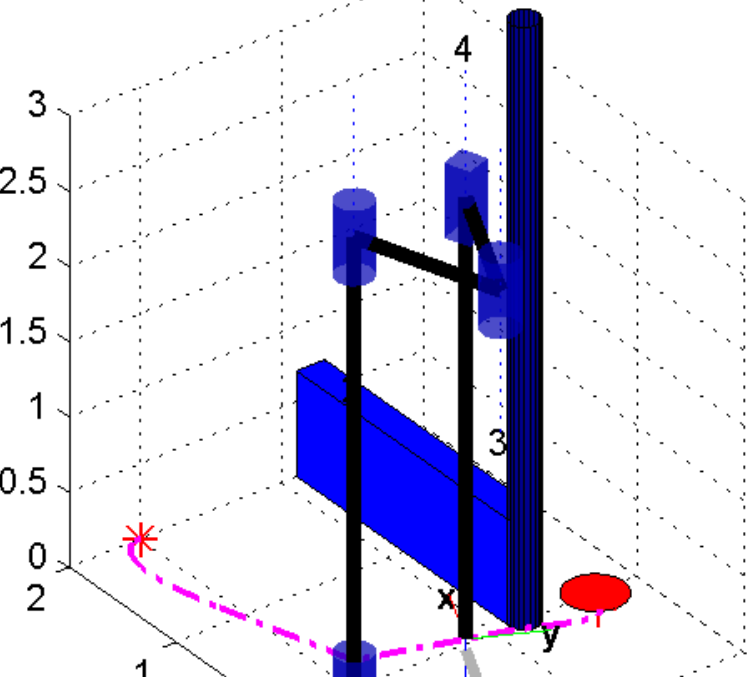}}
\subfigure[$t = 1.25$ sec]{
\includegraphics*[width=3.9cm,viewport=85 90 300 285]{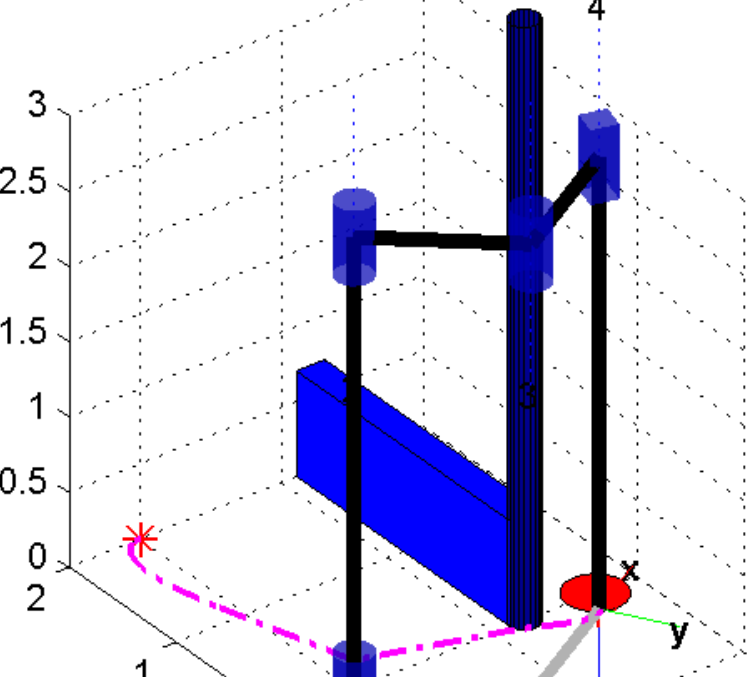}}
\caption{The motion of the SCARA robot corresponding to the resulting state trajectory when 5000 nodes are added to the tree with $R =  0.05\text{diag}(0.5, 0.5, 10)$.}
\label{fig:result4}
\end{figure}

\begin{figure}[t]
\vspace*{2mm}
\centering
\subfigure[Case 1, topview]{
\includegraphics*[width=3cm, height=3.8cm,viewport=108 35 330 300]{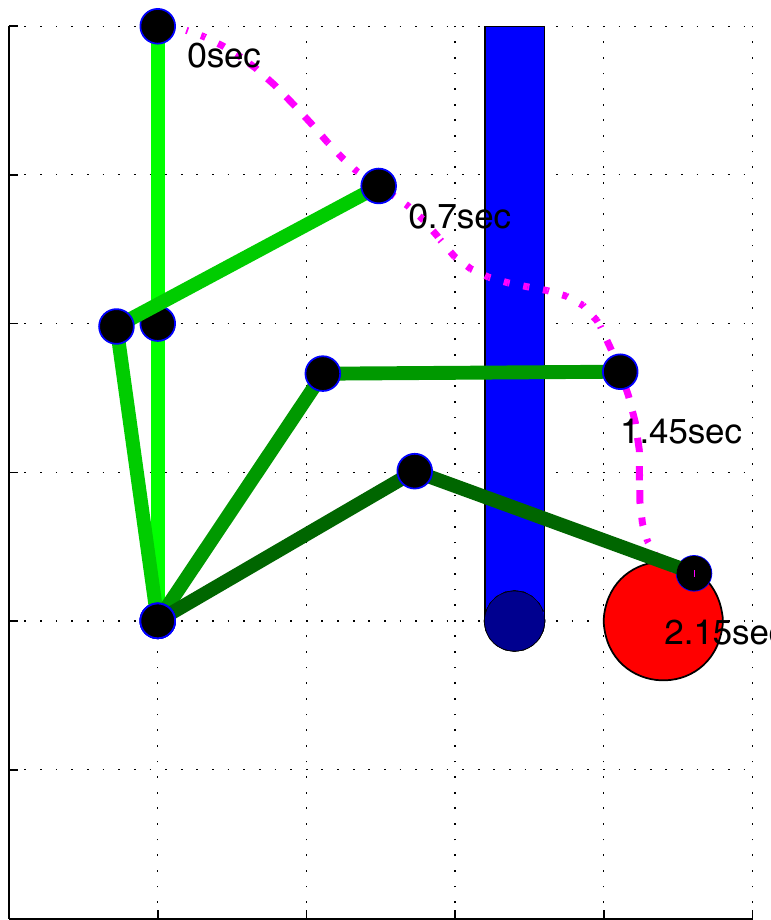}}
\subfigure[Case 1, input]{
\includegraphics*[width=5cm,viewport=20 10 400 300]{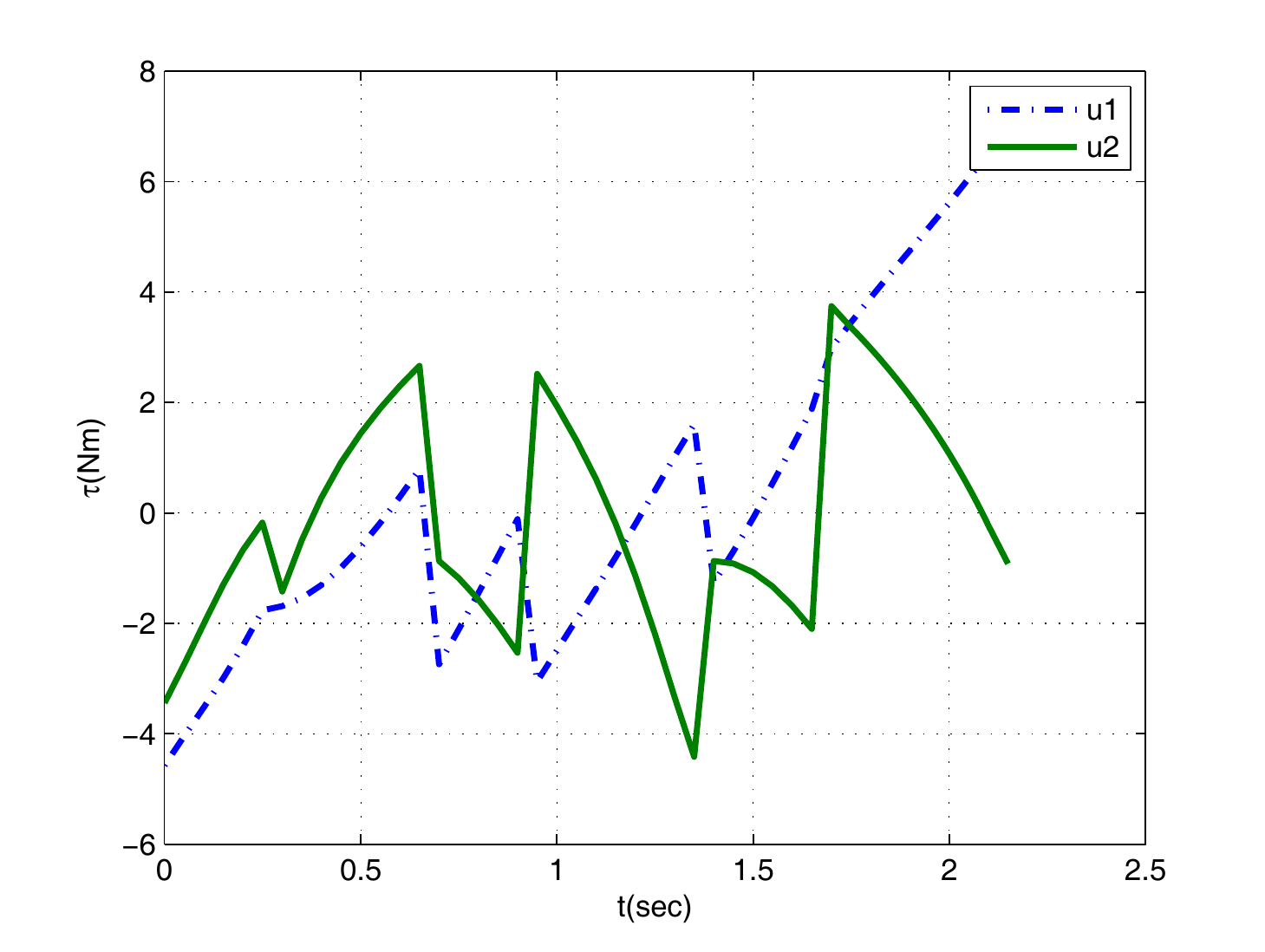}}
\subfigure[Case 2, topview]{
\includegraphics*[width=3cm, height=3.8cm,viewport=108 35 330 300]{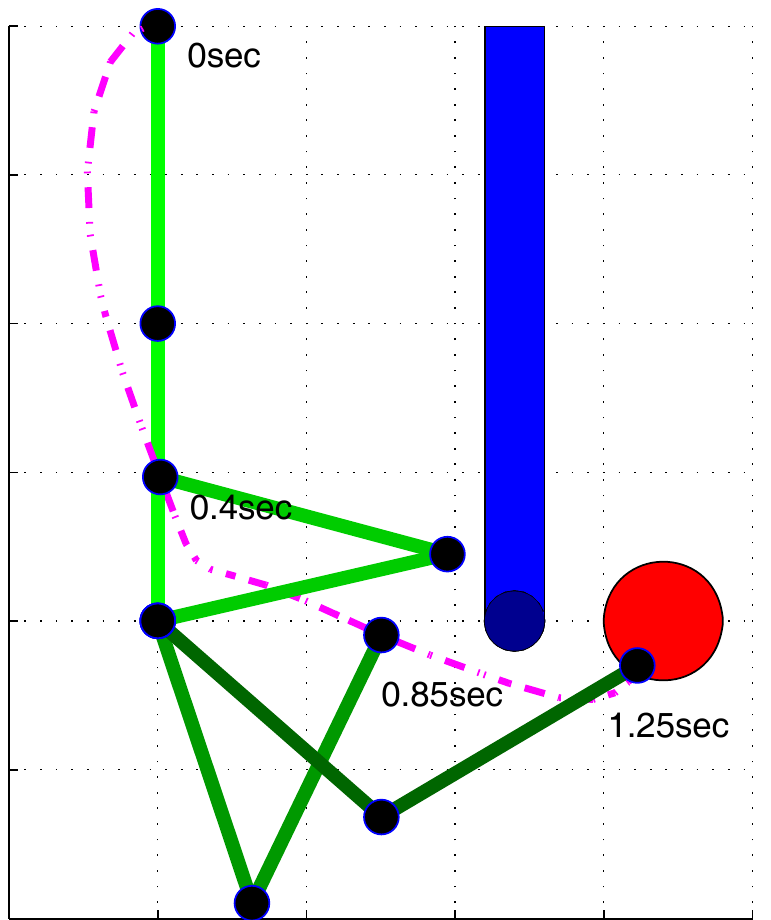}}
\subfigure[Case 2, input]{
\includegraphics*[width=5cm,viewport=20 10 400 300]{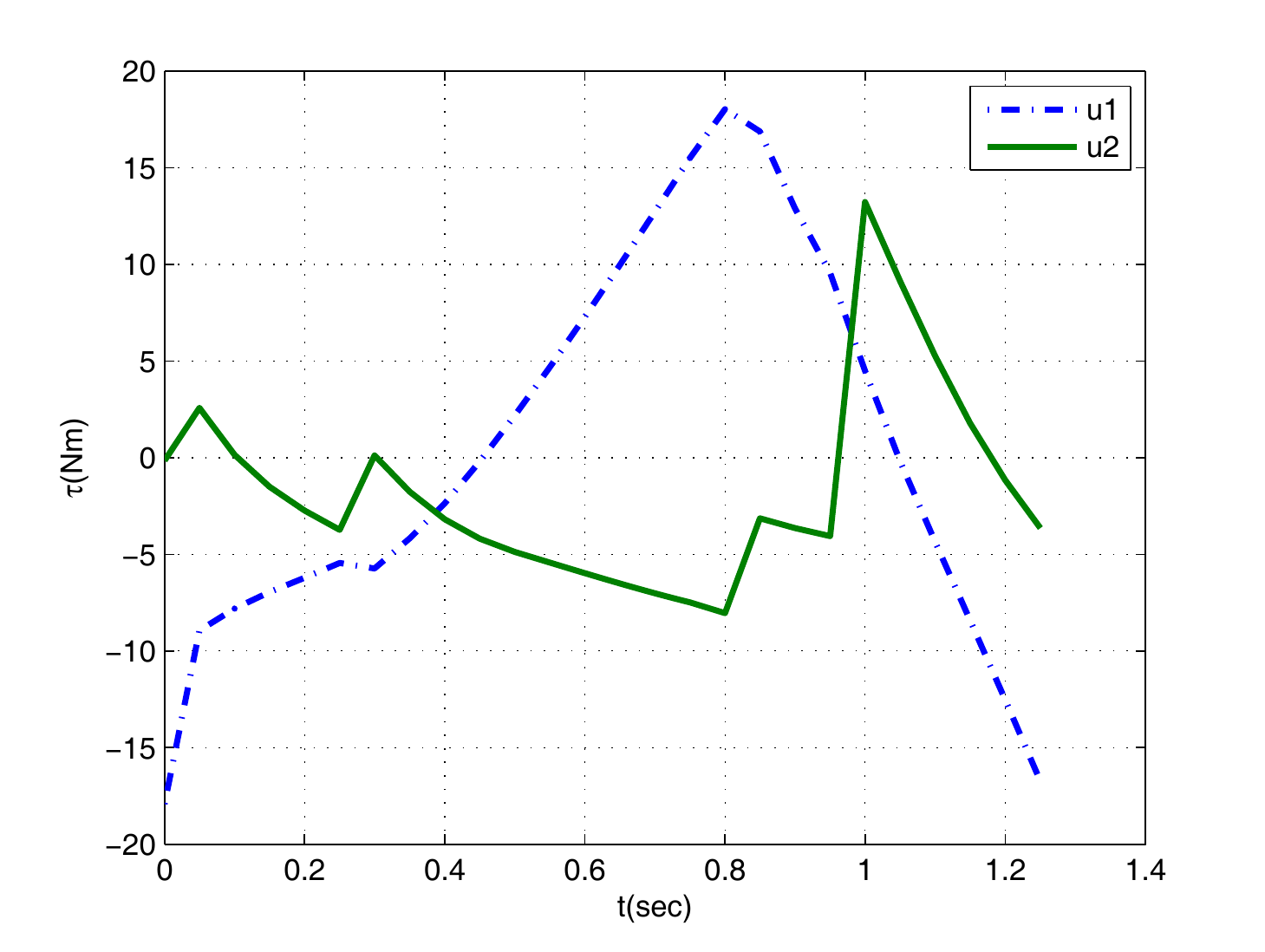}}
\caption{The top-views of resulting motion and the corresponding inputs.}
\label{fig:result5}
\end{figure}

The problem is to find the optimal motion trajectory that leads to the end-effector of the robot reaching the goal region (shown as red circle in Figs. \ref{fig:result3}-\ref{fig:result5}) from its initial state, $x(0) = [\pi/2~0~3~0~0~0]^T$ (the corresponding end-effector position is shown as red star in Figs \ref{fig:result3}--\ref{fig:result4}), while avoiding collision with obstacles.
The cost of the trajectory takes the form of  (\ref{eq:cost}).
With other parameters being fixed, the solutions are obtained with varying input penalty matrix, $R$, to see the effect of cost functional on the resulting trajectories.
Two values of $R$ are considered:
$$
R =  0.05\text{diag}(1, 1, 0.5),~0.05\text{diag}(0.5, 0.5, 10).
$$
The proposed VE-based method is implemented to solve nonlinear TPBVPs in the process of RRT*.
There are two homotopy classes for the solutions from the initial state to the final states:
the end effector 1) goes over the wall or 2) makes a detour around the wall.

Figs. \ref{fig:result3}--\ref{fig:result4} show the resulting motions of the robot when 5000 nodes are added into the tree for the two cases;
the magenta dashed line and the red circle represent the end-effector trajectory and the goal region, respectively.
For the first case ($R =  0.05\text{diag}(1, 1, 0.5)$), which has smaller control penalty on joint 3, the resulting trajectory goes over the wall as shown in Fig. \ref{fig:result3};
going over the wall by using cheap control input of joint 3 leads to lower cost in this case.
On the other hand, in the second case ($R =  0.05\text{diag}(0.5, 0.5, 10)$) shown in Fig. \ref{fig:result4}, the proposed algorithm generates the trajectory that makes a detour as the control penalty for the joint 3 is larger; going over the wall is so expensive in this case that the detouring trajectory is chosen as the best one.

Considering the control penalty for joint 1 and 2, the penalty is smaller in the second case than the first one; thus, the duration of the trajectory weighs more in the second case.
Fig. \ref{fig:result5} shows that the top-views of the SCARA robot motion at each time step in Figs. \ref{fig:result3}--\ref{fig:result4} and the corresponding control inputs;
the motion starts at the position colored in light-green and ends at the position colored in dark-green.
It can be seen that the trajectory reaches the goal state more quickly in the second case despite it requires the large amount of input-energy of the joint 1 and 2 (the arrival time of the minimum-cost trajectory is $2.15$ and $1.25$ second, respectively).

\section{Concluding Remarks}
In this work, an extension of RRT* to handle nonlinear kinodynamic differential constraints was proposed. In order to tackle two caveats: choosing a valid distance metric and solving two point boundary value problems to compute an optimal trajectory segment. An affine quadratic regulator (AQR)-based pseudo metric was adopted, and two iterative methods were proposed, respectively. The proposed methods have been tested on three numerical examples, highlighting their capability of generating dynamically feasible trajectories in various settings.

Despite the proposed methods have been focused on implementing to the RRT* framework, it is expected to help other state-of-the-art sampling-based motion planning algorithms, like RRT\# \cite{arslan2013use}, GR-FMTs \cite{Jeon2015optimal} or FMT* \cite{janson2015fast} to address nonlinear dynamical systems.

\ifCLASSOPTIONcaptionsoff
  \newpage
\fi

\bibliographystyle{IEEEtran}
\bibliography{IET_RRT}

\end{document}